\documentclass[11pt, a4paper, onecolumn, copyright]{AweAI}

\usepackage{multirow}
\usepackage[authoryear, sort&compress, round]{natbib}
\usepackage{graphicx}
\usepackage{xspace}
\usepackage{microtype}
\usepackage{subcaption}
\usepackage{booktabs}

\usepackage{listings} 
\usepackage[most]{tcolorbox}
\tcbuselibrary{listings}

\usepackage{textcomp}
\usepackage{hyperref}
\usepackage{amsmath}
\usepackage{amssymb}
\usepackage{mathtools}
\usepackage{amsthm}
\usepackage{enumitem}

\usepackage{pifont}
\definecolor{darkgreen}{RGB}{50,100,0}
\definecolor{darkred}{RGB}{200, 0, 0}
\newcommand{\cmark}{\textcolor{darkgreen}{\ding{51}}} %
\newcommand{\xmark}{\textcolor{darkred}{\ding{55}}}

\usepackage{xurl}
\usepackage{titletoc}

\definecolor{tableblue}{HTML}{F2F7FF}
\definecolor{headergray}{HTML}{444444}

\colorlet{DarkGreen}{green!50!black}
\colorlet{DarkRed}{red}

\newcommand{\DeltaNote}[1]{%
  ~\rlap{\scriptsize\textcolor{DarkRed}{{#1}}}%
}
\newcommand{\DeltaNotePlus}[1]{%
  ~\rlap{\scriptsize\textcolor{DarkGreen}{{#1}}}%
}

\DeclareRobustCommand{\github}{\raisebox{-1.5pt}{\includegraphics[height=1.05em]{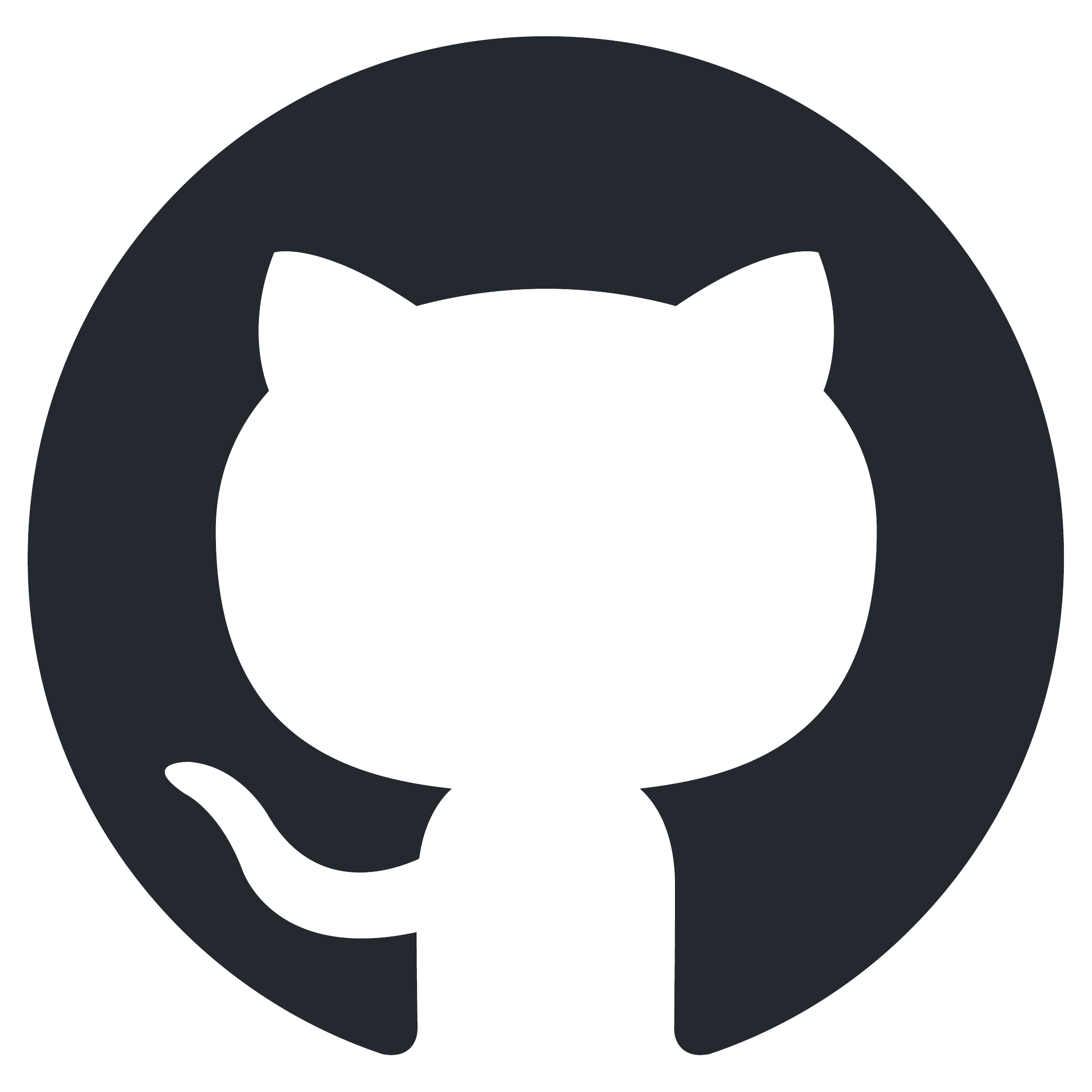}}\xspace}
\DeclareRobustCommand{\huggingface}{\raisebox{-1.5pt}{\includegraphics[height=1.05em]{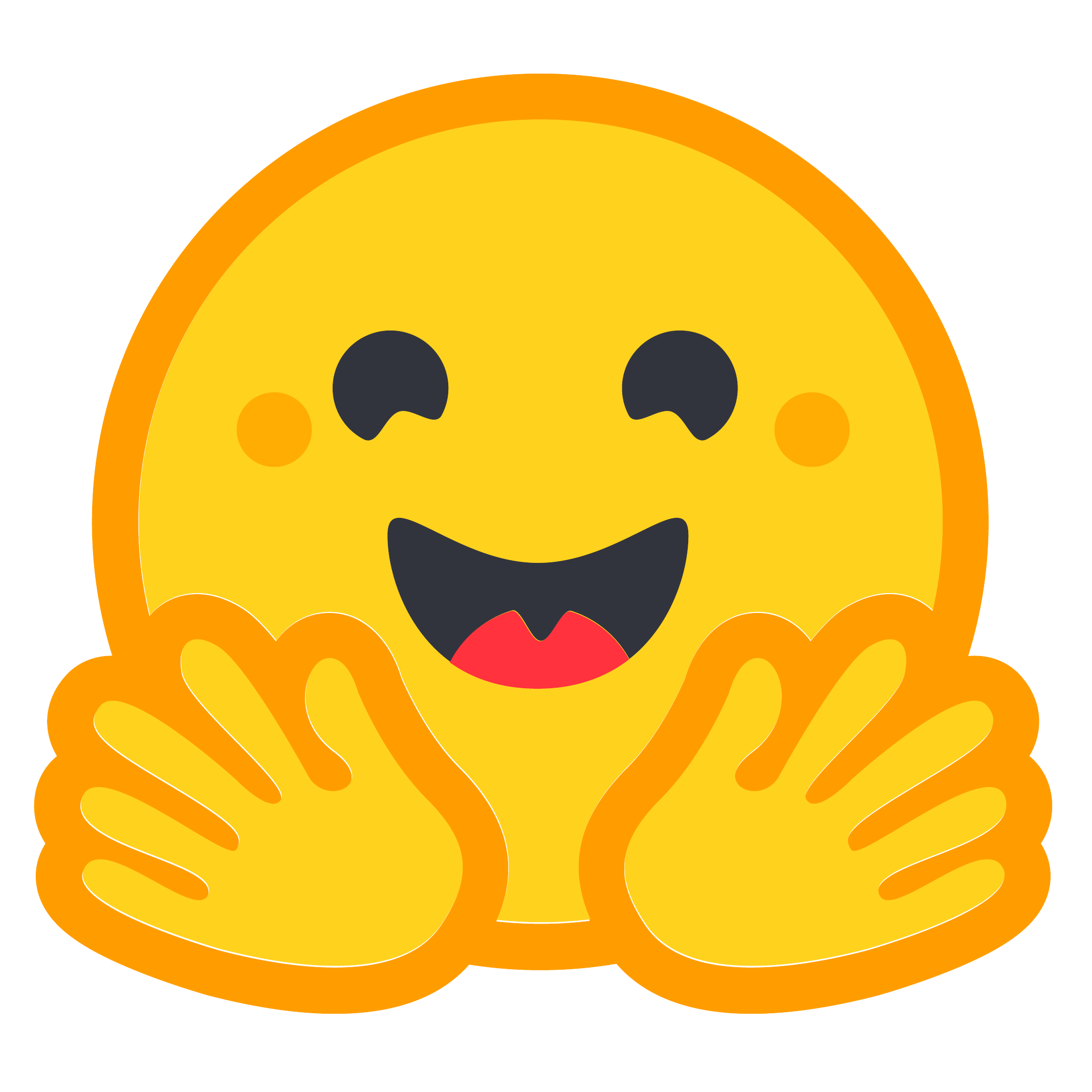}}\xspace}

\DeclareRobustCommand{\webpage}{\raisebox{-1.5pt}{\includegraphics[height=1.05em]{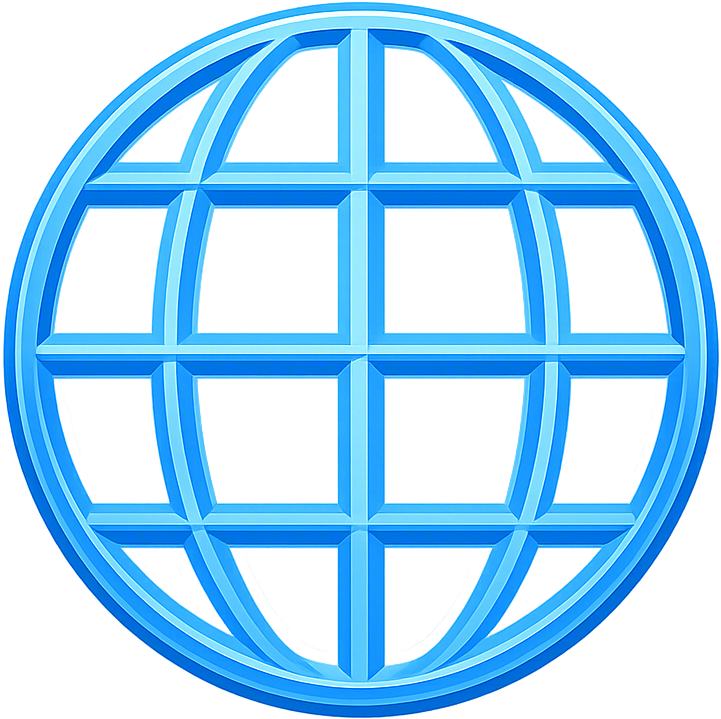}}\xspace}
\DeclareRobustCommand{\scaffold}{\raisebox{-1.5pt}{\includegraphics[height=1.05em]{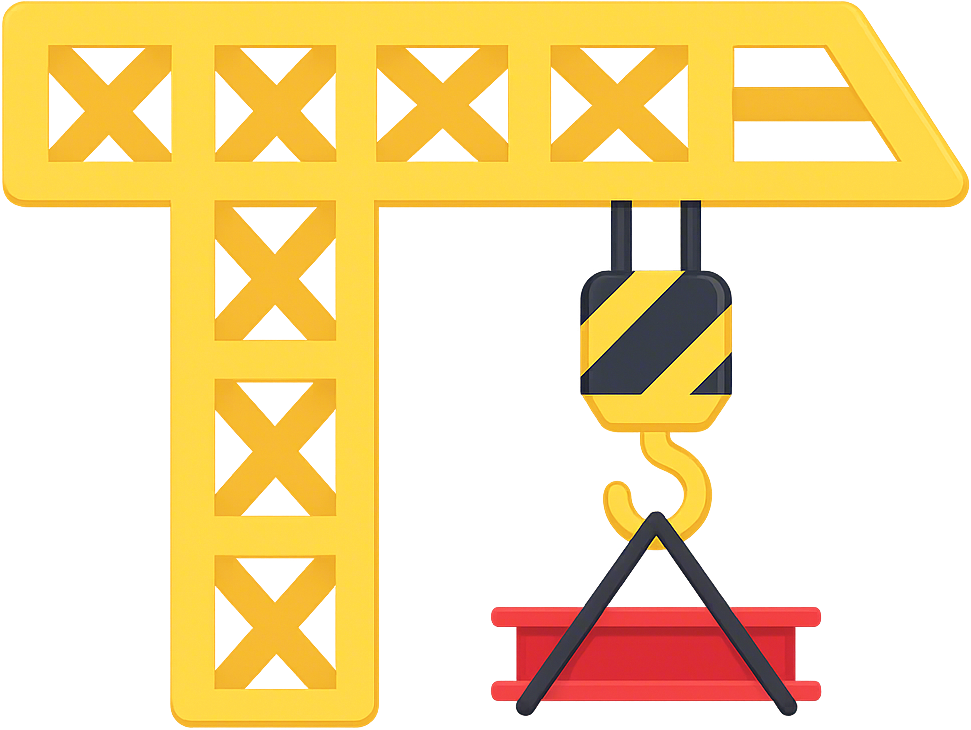}}\xspace}

\newcommand{\footerlinks}{%
  \huggingface\ \href{https://huggingface.co/datasets/AweAI-Team/BeyondSWE}{Benchmark}\hspace{1.5em}%
  \github\ \href{https://github.com/AweAI-Team/BeyondSWE}{Repo}\hspace{1.5em}
  \scaffold\ \href{https://github.com/AweAI-Team/AweAgent}{Scaffold}\hspace{1.5em}
  \webpage\ \href{https://aweai-team.github.io/BeyondSWE/}{WebPage}%
}

\newcommand{\papername}{BeyondSWE}
\newcommand{\benchmark}{BeyondSWE}
\newcommand{\agent}{SearchSWE}

\newtcblisting{markdownbox}[1][]{
  listing engine=listings,
  listing options={
    basicstyle=\scriptsize\ttfamily, 
    breaklines=true,              
    breakatwhitespace=true,
    numbers=none,                 
    columns=fullflexible,         
    keepspaces=true,              
    aboveskip=0pt,
    belowskip=0pt
  },
  colback=gray!5!white,           
  colframe=gray!70!black,         
  title={\textbf{Prompt (Markdown)}}, 
  fonttitle=\bfseries,
  listing only,                   
  left=5mm, right=5mm, top=3mm, bottom=3mm, 
  breakable,                      
  enhanced,                       
  overlay={\begin{tcbclipinterior}\fill[gray!5!white] (frame.south west) rectangle ([xshift=0mm]frame.north west);\end{tcbclipinterior}}, 
  #1                              
}

\newtcbinputlisting{\markdownfile}[2][]{
  listing engine=listings,
  listing file={#2}, 
  listing options={
    basicstyle=\scriptsize\ttfamily,
    breaklines=true,
    breakatwhitespace=true,
    numbers=none,
    columns=fullflexible,
    keepspaces=true
  },
  colback=gray!5!white,
  colframe=gray!70!black,
  title={\textbf{Prompt (Markdown)}},
  fonttitle=\bfseries,
  listing only,
  breakable,
  enhanced,
  #1
}

\definecolor{lessonBg}{HTML}{EEF6FF}
\definecolor{lessonRule}{HTML}{2F6FD6}

\newenvironment{lessonbox}{%
  \par\smallskip
  \begin{mdframed}[
    linecolor=lessonBg,
    linewidth=0pt,
    roundcorner=5pt,
    backgroundcolor=lessonBg,
    innerleftmargin=7pt,
    innerrightmargin=7pt,
    innertopmargin=5pt,
    innerbottommargin=5pt,
    skipabove=5pt,
    skipbelow=6pt
  ]%
  \small\noindent
  \textcolor{lessonRule}{\textbf{\textit{Lesson.}}}%
}{%
  \end{mdframed}%
  \par\smallskip
}

\definecolor{exampleBg}{HTML}{F3F7FB}
\definecolor{exampleRule}{HTML}{7A9BBE}
\definecolor{promptBg}{HTML}{F8F8F8}
\definecolor{promptRule}{HTML}{666666}

\newtcolorbox{examplebox}[2][]{
  enhanced,
  colback=exampleBg,
  colframe=exampleRule,
  boxrule=0.5pt,
  arc=3pt,
  left=6pt,
  right=6pt,
  top=5pt,
  bottom=5pt,
  fonttitle=\bfseries,
  coltitle=white,
  colbacktitle=exampleRule,
  title={#2},
  before upper={\setlength{\parskip}{0.45\baselineskip}\setlength{\parindent}{0pt}\normalfont},
  #1
}

\newtcblisting{promptbox}[2][]{
  enhanced, breakable,
  colback=promptBg, colframe=promptRule,
  boxrule=0.4pt, arc=2pt,
  left=5pt, right=5pt, top=4pt, bottom=4pt,
  fonttitle=\bfseries, coltitle=white,
  colbacktitle=promptRule,
  title={#2},
  listing only,
  listing options={
    basicstyle=\ttfamily\scriptsize,
    breaklines=true,
    columns=fullflexible,
    keepspaces=true,
    showstringspaces=false
  },
  #1
}

\uselogo{}

\title{\papername: Can Current Code Agent Survive Beyond Single-Repo Bug Fixing?}

\renewcommand{\today}{}
\newcommand{\publicday}{Feb.~18, 2026}

\author[*]{Guoxin Chen}
\author[*]{Fanzhe Meng}
\author[*]{Jiale Zhao}

\author[ \hspace{-0.3em}]{Minghao Li}
\author[ \hspace{-0.3em}]{Daixuan Cheng}
\author[ \hspace{-0.3em}]{Huatong Song}
\author[ \hspace{-0.3em}]{Jie Chen}
\author[ \hspace{-0.3em}]{Yuzhi Lin}
\author[ \hspace{-0.3em}]{Hui Chen}
\author[ \hspace{-0.3em}$^\dag$]{Xin Zhao}
\author[ \hspace{-0.3em}$^\dag$]{Ruihua Song}
\author[ \hspace{-0.3em}]{Chang Liu}
\author[ \hspace{-0.3em}]{Cheng Chen}
\author[ \hspace{-0.3em}$^\dag$]{Kai Jia}
\author[ \hspace{-0.3em}]{Ji-Rong Wen}

\correspondingauthor{\{gx.chen.chn, mengfanzhe16, marshmallowzjl, batmanfly, jiakai0419\}@gmail.com, songruihua\_bloon@outlook.com}

\affil[1]{Gaoling School of Artificial Intelligence, Renmin University of China}
\affil[2]{Independent Researcher}
\affil[3]{AweAI Team\footnote{$^*$Equal Contributions. $^\dag$Corresponding authors.\hfill\textbf{Date:} \publicday.}}

\begin{abstract}
Current code-agent benchmarks primarily evaluate localized issue resolution within a single target repository, leaving under-tested many software engineering tasks that require external knowledge or broader repository-level changes.
We introduce \benchmark{}, a 500-instance benchmark drawn from 246 real-world GitHub repositories to evaluate code agents beyond single-repository bug fixing.
\benchmark{} covers four representative settings: cross-repository issue resolution, domain-specific issue resolution, dependency-driven migration, and document-to-repository generation, spanning both broader knowledge scope and broader resolution scope.
Our evaluation shows that \benchmark{} remains far from saturated: the best OpenHands-based agent reaches 46.12 average score, while the strongest Codex harness with GPT-5.4 (xhigh) reaches 56.65 under a search-aware prompt.
To study whether external information access closes this gap, we use \agent{} as a controlled diagnostic baseline for search-augmented coding.
Search access improves most models and substantially helps some tasks, but the gains remain limited and uneven, showing that current agents still struggle to convert retrieved information into precise, version-compatible, and locally actionable code changes.
These results suggest that deep search for coding remains an open problem: progress requires agents that can reliably combine external evidence with repository-local reasoning and execution-based verification.

\centerline{\footerlinks}
\end{abstract}

\makeatletter
\begin{document}

\begingroup
\makeatletter
\renewcommand{\thefootnote}{}
\renewcommand{\@makefnmark}{}
\maketitle
\makeatother
\endgroup

\section{Introduction}
Modern code agents have made rapid progress on software-engineering benchmarks, led by SWE-bench Verified~\citep{jimenez2024swebench}, which grounds evaluation in real GitHub issues.
Subsequent work has extended this paradigm through live updates, broader repository coverage, multilingual settings, and more complex issue instances~\citep{zhang2025swebench,zan2025multiswebench,deng2025swe}.

Yet this evaluation paradigm still leaves important parts of software engineering under-tested. 
Most existing benchmarks center on localized issue resolution within a single target repository, where the relevant context is expected to be recoverable from the issue and the codebase.
In practice, many engineering tasks fail precisely because the necessary information is outside the repository, or because the required change must be coordinated across a broader portion of the system. 
Developers may need to consult upstream projects, documentation, or domain knowledge.
They may also need to migrate APIs across a codebase or build a coherent repository from a specification.
These settings stress capabilities that are only partially captured by single-repository bug-fixing benchmarks.

This gap motivates a simple question: \emph{how well do current code agents perform beyond single-repository bug fixing?} 
To study this question, we introduce \textbf{\benchmark{}}, a benchmark designed around two practical dimensions.
The first, \emph{knowledge scope}, asks whether solving a task requires only repository-local information or also external software, domain, documentation, or specification knowledge.
The second, \emph{resolution scope}, asks whether the required solution is a localized fix, a repository-wide transformation, or full repository construction.
Together, these dimensions allow us to stress-test capabilities that remain underrepresented in existing benchmarks while retaining executable, test-based evaluation.

Following this lens, we define four representative stress tests.
\emph{Cross-repository issue resolution (CrossRepo)} keeps the familiar issue-resolution format but requires agents to use information from related external repositories.
\emph{Domain-specific issue resolution (DomainFix)} tests whether agents can combine code reasoning with specialized knowledge from scientific and engineering domains. 
\emph{Dependency-driven migration (DepMigrate)} evaluates repository-wide coordination under breaking changes in upstream dependencies. 
\emph{Document-to-repository generation (Doc2Repo)} asks agents to construct a functional repository from a natural-language specification. 
These tasks comprise 500 instances drawn from 246 real-world GitHub repositories.
Their target solutions affect an average of 10.9 files and 1039 lines per instance, substantially exceeding the modification scale of existing SWE-bench-style benchmarks.
Our evaluation shows that \benchmark{} remains challenging even for strong coding agents. Under the OpenHands scaffold~\citep{wang2025openhands}, the best model reaches 46.12 average score, far from saturating the benchmark.
The results expose task-specific weaknesses: CrossRepo stresses external software knowledge, DomainFix requires domain-specific reasoning, DepMigrate demands coordinated repository-level edits, and Doc2Repo often yields partial functionality without fully correct repositories.

\begin{table}[t]
\centering
\resizebox{0.6\linewidth}{!}{
\begin{tabular}{@{}lccccc@{}}
\toprule
\multirow{2}{*}{\textbf{Benchmark}} & \multicolumn{2}{c}{\textbf{Scope}}   & \multicolumn{3}{c}{\textbf{Statistics}}                \\ \cmidrule(lr){2-3} \cmidrule(lr){4-6} 
                                    & \textbf{Resol.} & \textbf{Knowledge} & \textbf{\#Repo} & \textbf{\#Files} & \textbf{\#Lines}  \\ \midrule
SWE-bench-Verified                        & Local Func      & Within Repo        & 12              & 1.3              & 11.6              \\
SWE-bench-Live                            & Local Func      & Within Repo        & \underline{223} & 2.7              & 65.1              \\
SWE-bench Pro                             & Local Func      & Within Repo        & 41              & \underline{4.1}  & \underline{107.4} \\ \midrule
\rowcolor{CornflowerBlue!15}CrossRepo                           & Local Func      & Cross Repo         & 67              & 4.1              & 190.7             \\
\rowcolor{CornflowerBlue!15}DomainFix                           & Local Func      & Domain             & 12              & 4.2              & 157.6             \\
\rowcolor{CornflowerBlue!15}DepMigrate                          & Global Repo     & Official Docs      & 120             & 8.4              & 281.6             \\
\rowcolor{CornflowerBlue!15}Doc2Repo                            & Global Repo     & Human Spec         & 50              & 26.8             & 3528.4            \\ \midrule
\rowcolor{CornflowerBlue!15}
\textbf{\benchmark{}}               & \textbf{Mix}    & \textbf{Mix}       & \textbf{246}    & \textbf{10.9}     & \textbf{1039.6}    \\ \bottomrule
\end{tabular}
}
\caption{Comparison with existing SWE benchmarks}
\label{tab:bench_statistics}
\end{table}

Because several tasks require information beyond the local repository, we instantiate \textbf{\agent{}} as a controlled diagnostic baseline for search-augmented coding.
\agent{} minimally extends a code-agent workflow with web search and fetch tools, allowing us to compare otherwise similar settings with and without external information access.
The results indicate that search access for coding is genuinely useful: most models improve under \agent{}, and the Codex harness using GPT-5.4 (xhigh) improves from 48.48 to 56.65 when explicitly prompted to integrate search into the coding workflow.
Yet these gains remain limited and uneven, and our case studies show that agents often fail to turn retrieved information into precise, version-compatible, and locally actionable code changes.
This suggests that deep search for coding remains an open problem: search and coding have each advanced rapidly, but current agents do not yet reliably synthesize them into robust search-augmented coding workflows.

To summarize, our contributions are as follows:
\begin{itemize}[topsep=1pt, partopsep=1pt, leftmargin=*, itemsep=-1pt]
\item We introduce \benchmark{}, a 500-instance benchmark for evaluating code agents beyond single-repository bug fixing, spanning four representative settings that vary in knowledge scope and resolution scope.
\item We conduct a broad empirical evaluation of frontier and code-specialized agents, showing that current systems remain brittle across cross-repository reasoning, domain-specific repair, dependency migration, and repository generation.
\item We use \agent{} as a controlled diagnostic baseline to study deep search for coding, showing that search access is useful but insufficient: current agents still struggle to convert retrieved information into precise, version-compatible, and locally actionable code changes. 
\end{itemize}

\section{Related Work}
\label{sec:related_work}
\textbf{SWE Benchmark.}
SWE-bench-Verified~\citep{chowdhury2024introducing} established executable GitHub-issue resolution as the dominant evaluation setting for code agents.
Follow-up benchmarks improve this setting through live updates, multilingual coverage, decontamination, and harder issue instances~\citep{tian2024scicode,liu2025e2edev,liu2025migrationbench,yang2025swebench,zan2025multi,rashid2025swe,badertdinov2025swerebench,ding2025nl2repo,amin2026jmigbench}.
Most preserve the core assumption that the target repository contains the main problem-solving context.
\benchmark{} keeps executable test-based evaluation, but shifts the task distribution toward external knowledge requirements and broader resolution scopes.

\textbf{Code agents and search.}
Recent frontier coding harnesses, including Codex and Claude Code~\citep{openai2025codexupgrades,anthropic2025websearchapi}, increasingly expose web search or fetch capabilities as part of coding workflows.
However, these production systems couple search with proprietary prompts, tool policies, model choices, and scaffolding, making it difficult to isolate how external information access affects coding performance.
We therefore instantiate \agent{} as a controlled search-augmented baseline, allowing us to study when search helps, when it remains insufficient, and why on \benchmark{}.
\section{\benchmark{}}
\label{sec:bench}
\begin{figure*}[t]
    \centering
    \includegraphics[width=\linewidth]{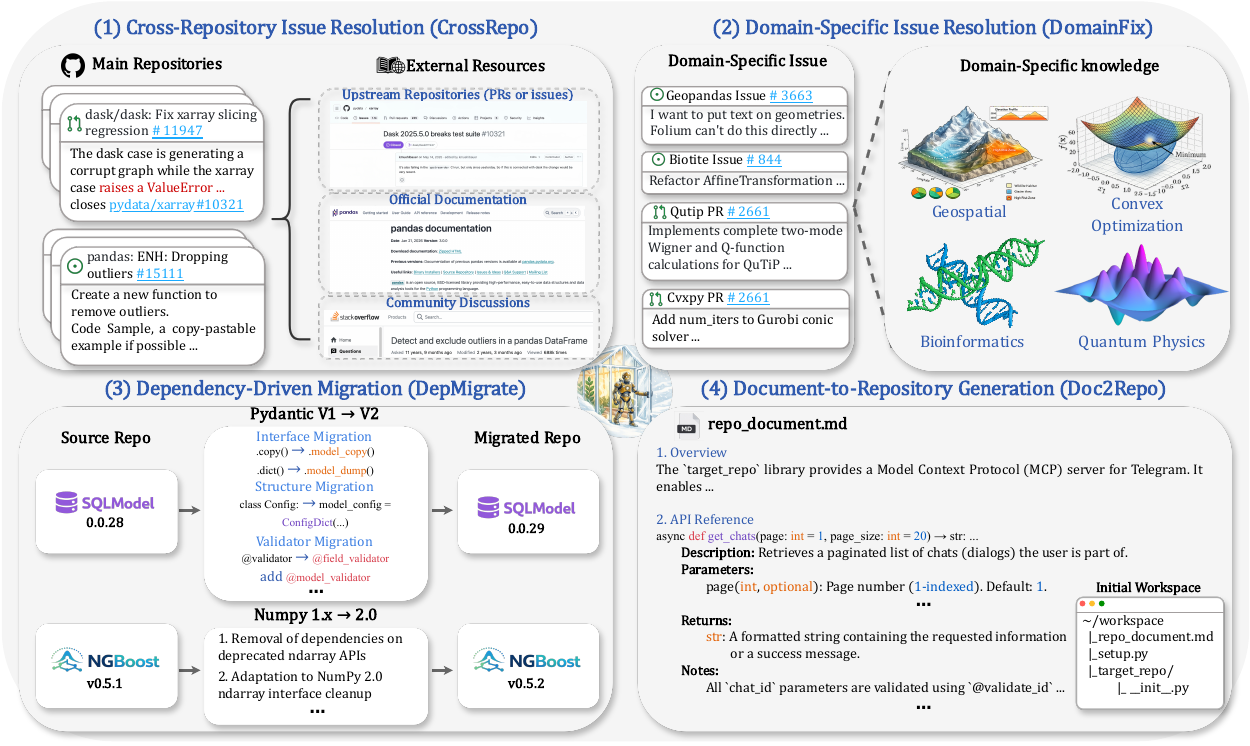}
    \caption{Overview of \benchmark{}. Our benchmark extends evaluation along two practical dimensions—\emph{knowledge scope} and \emph{resolution scope}: CrossRepo and DomainFix expand knowledge scope by requiring external software resources and domain expertise respectively; DepMigrate and Doc2Repo expand resolution scope from localized patches to codebase-wide transformations.}
    \label{fig:benchmark}
\end{figure*}

\benchmark{} is a benchmark for evaluating code agents on software engineering tasks that require broader knowledge or broader resolution scope than single-repository bug fixing.
As shown in Figure~\ref{fig:benchmark}, we use two practical dimensions to guide task selection.
\emph{Knowledge scope} distinguishes tasks whose relevant information is recoverable from the target repository from tasks that require external software artifacts, scientific domain knowledge, API documentation, or a specification.
\emph{Resolution scope} distinguishes localized fixes from repository-wide transformations and full repository construction.
These dimensions serve as a design lens for stress-testing underrepresented capabilities while preserving executable, test-based evaluation.
Overall, \benchmark{} contains 500 instances drawn from 246 GitHub repositories.
The target solutions affect an average of 10.9 files and 1039.6 lines per instance, substantially exceeding the modification scale of existing SWE-bench-style benchmarks.

\subsection{Task Formulation}
Each instance contains three components.
First, the \emph{problem statement} specifies the task to be solved.
For CrossRepo, DomainFix, and DepMigrate, it follows the GitHub issue-resolution format.
For Doc2Repo, it is a natural-language specification of the target repository's API and behavior.
Second, the \emph{Docker environment} provides a reproducible runtime with the target repository, dependencies, and test commands.
Third, the \emph{test suite} defines executable success criteria.
For CrossRepo, DomainFix, and DepMigrate, we follow the SWE-bench convention of pass-to-pass (P2P) tests that should remain passing and fail-to-pass (F2P) tests that should pass only after the correct fix is applied.
For Doc2Repo, where the agent starts from an empty workspace, success is measured by the complete test suite for the generated repository.

\subsection{Benchmark Tasks}

\textbf{Cross-repository issue resolution.}
CrossRepo keeps the familiar issue-resolution setting but requires agents to use information from related external repositories or linked artifacts.
This setting reflects cases where a bug report, API behavior, or implementation pattern cannot be understood from the target repository alone.
To construct the task, we scan Python-dominant GitHub repositories for merged pull requests containing external links and collect about 3,000 candidates.
After environment construction and stability filtering, about 800 candidates remain.
We manually verify that the external links are relevant to the issue and that the rewritten problem statement preserves the task context without revealing solution-specific details.
This process yields 200 issues across 67 repositories, with an average of 1.3 external links per issue.

\textbf{Domain-specific issue resolution.}
DomainFix evaluates whether agents can combine code reasoning with specialized knowledge from scientific and engineering domains.
The target projects come from 11 research fields, including quantum physics, molecular dynamics, geospatial analysis, bioinformatics, and materials science.
We collaborate with domain experts to identify 21 high-quality repositories and collect about 800 candidate pull requests.
After executable environment construction, around 200 candidates pass stability inspection.
Each remaining instance is independently reviewed by three domain experts for environmental correctness, genuine domain complexity, and solution non-triviality.
Only unanimously accepted instances are retained.
The final task contains 72 issues across 12 repositories.

\textbf{Dependency-driven migration.}
DepMigrate tests whether agents can coordinate repository-level edits under breaking changes in upstream dependencies.
Unlike localized bug fixing, migration requires agents to understand an upstream API change, find affected call sites, and update the codebase consistently.
We identify 23 widely used packages with significant version upgrades and collect pull requests whose descriptions or commit messages mention these packages and relevant version numbers.
LLM-based filtering removes candidates that are not genuine migration efforts, producing about 7,000 candidates before environment construction.
After stability inspection, about 1,000 candidates remain.
Four software engineering experts then verify migration validity, producing 178 issues across 120 repositories.
For each instance, the environment installs the upgraded dependency while the codebase is checked out before the migration patch, so agents must adapt the repository to the new dependency behavior.

\textbf{Document-to-repository generation.}
Doc2Repo evaluates repository construction from a specification rather than repair of an existing codebase.
Agents receive a natural-language document describing the intended API and behavior, and must create a complete repository from an empty workspace.
To reduce contamination risk, we collect Python repositories created between January and November 2025, require continued activity after August 2025, and retain projects with at least three contributors and more than 20 stars.
For each repository, Gemini 3 Pro~\citep{googledeepmind2026gemini3pro} explores the codebase and generates a specification covering purpose, usage examples, public classes and functions, parameters, return types, and behavior, while removing implementation details and directory structure.
We mask repository names with \texttt{target\_repo} and require agents to infer the structure from contextual cues such as import paths.
We adapt tests from the original repositories with LLM assistance and human review.
After environment construction, 60 candidates remain, from which we select 50 high-quality instances.

\begin{figure*}[t]
    \centering
    \includegraphics[width=\linewidth]{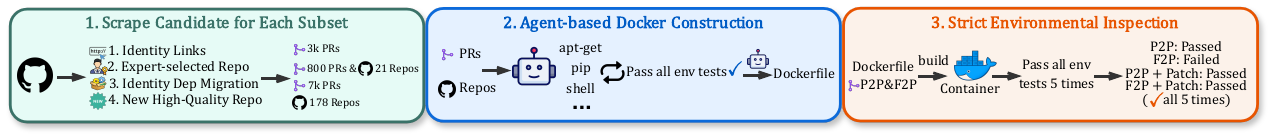}
    \caption{Environment Construction Pipeline for \benchmark{}. We collect task-specific candidates, build executable Docker environments through an agent-assisted setup process, and retain only instances whose tests exhibit stable fail-to-pass and pass-to-pass behavior across repeated runs.}
    \label{fig:docker}
\end{figure*}

\subsection{Environment Construction}
Reliable evaluation requires runnable historical environments, but many real repositories suffer from dependency decay, unavailable packages, deprecated APIs, and missing system libraries.
As shown in Figure~\ref{fig:docker}, we use an agent-assisted process to construct each Docker environment.
The agent starts from a base Ubuntu container, clones the repository, checks out the pre-PR commit, and iteratively resolves setup failures until the existing tests can be executed.
The agent can use shell commands to install missing packages, compilers, and libraries, which are often absent from repository-level dependency files.
We then distill the successful command history into a reproducible \texttt{Dockerfile}.

Each generated environment undergoes strict stability inspection.
We build the Docker image and execute the relevant tests five times.
For CrossRepo, DomainFix, and DepMigrate, we require P2P tests to pass and F2P tests to fail before applying the reference patch, and require both sets to pass after applying the patch.
Instances with flaky behavior, incomplete setup, or invalid fail-to-pass transitions are discarded.
For Doc2Repo, the environment is validated by running the adapted test suite against the reference implementation and by manually auditing tests whose expected behavior depends on the generated specification.

\subsection{Evaluation Protocol}
\label{sec:evaluation}
We separate agent execution from final verification so that scores reflect the submitted code changes rather than artifacts of the agent's workspace.
During evaluation, an agent works in its own Docker environment and produces a patch or a generated repository.
We then extract the resulting changes and apply them to a fresh container that is independent of the agent's workspace.
This prevents environmental side effects, cached artifacts, or modified local configuration from affecting the final score.

We also apply integrity safeguards against solution leakage and test manipulation.
Following concerns raised in prior benchmark analyses~\citep{xiao2026mimo}, we remove git commits, logs, and metadata after the target commit while retaining earlier history that a developer could realistically inspect.
After applying an agent patch, we restore all test files to their original state before running evaluation.
Thus, success must come from changing the target implementation rather than editing tests or exploiting future repository history.

For CrossRepo, DomainFix, and DepMigrate, we report \textbf{Resolved Rate}, the percentage of instances where all P2P and F2P tests pass in the fresh container.
For Doc2Repo, we report \textbf{Pass Rate}, the average percentage of tests passed, and \textbf{(Almost) Correct Count}, the number of repositories that pass all tests or at least 90\% of tests.

\subsection{Quality Control and Human Verification}
\label{sec:quality_control}
\benchmark{} combines automated filtering with task-specific human verification.
Automated filtering removes candidates with unstable environments, invalid test transitions, flaky behavior, or solution-revealing problem statements.
Human review checks whether retained tasks represent genuine SWE challenges rather than artifacts of environment setup or underspecified tests.

The verification process uses task-appropriate expertise.
DomainFix uses domain experts to confirm that each retained issue requires specialized scientific or engineering knowledge beyond general programming.
DepMigrate uses software engineering experts to verify that each instance corresponds to a real dependency migration rather than an incidental version bump.
Across all tasks, five senior software engineers inspect environment construction and data cleaning, and five senior PhD researchers in software engineering and LLMs audit final task quality.
For Doc2Repo, tests are additionally reviewed to ensure that the specification and expected behavior are aligned; when a test reflects stricter behavior than the original repository, the requirement is documented in the instance metadata.
Detailed reviewer backgrounds, agreement criteria, rejection reasons, and compensation information are reported in the appendix.

\begin{figure}[t]
    \centering
    \includegraphics[width=0.7\linewidth]{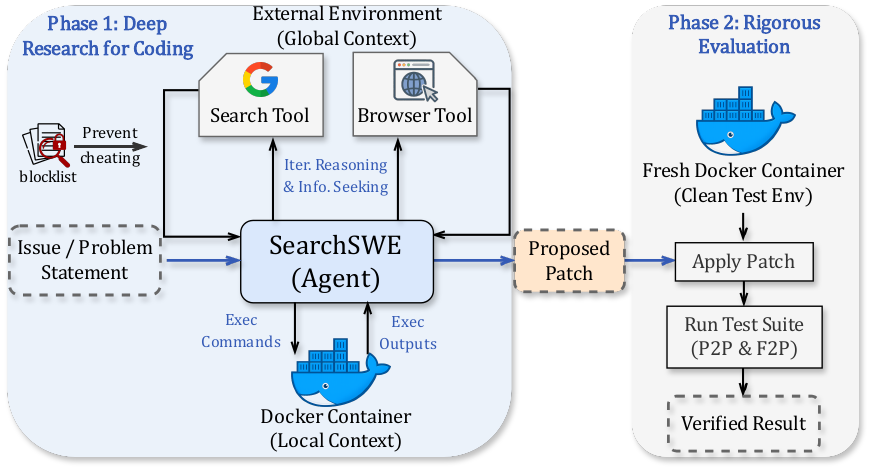}
    \caption{Overview of \agent{} as a controlled search-augmented coding baseline. The agent retains the local coding workflow, while adding web search and fetch tools under a blocklist that prevents direct access to solution-revealing target-repository artifacts.}
    \label{fig:searchswe}
\end{figure}

\section{Search-Augmented Diagnostic Baseline}
\label{sec:searchswe}
Several \benchmark{} tasks require information that may not be present in the target repository.
CrossRepo may depend on related implementations or upstream discussions, DomainFix may require specialized scientific knowledge, and DepMigrate may require migration guides or dependency documentation.
This motivates a controlled diagnostic question: \textit{when a code agent has access to external information, does it reliably turn that information into correct code changes?}

We instantiate \agent{} to study this question.
\agent{} minimally extends a standard code-agent workflow~\citep{wang2025openhands} with two external-information tools: a \emph{search tool} that queries the web for potentially relevant resources, and a \emph{browser tool} that fetches and summarizes the content of a specified webpage.
This design lets us compare otherwise similar settings with and without external information access, isolating the effect of search augmentation on coding performance.

\textbf{Controlled access.}
To keep the comparison focused, \agent{} exposes web search and fetch as general tools rather than task-specific retrieval modules.
The agent decides when to search, what query to issue, which result to inspect, and how to integrate retrieved evidence into local implementation and verification.
This setup is deliberately simple: performance improvements are informative, but the primary goal is to diagnose whether current agents can combine search with coding under realistic repository constraints.

\textbf{Cheating prevention.}
Because \benchmark{} instances are derived from real repositories, unrestricted web access could allow an agent to retrieve the original issue or pull request.
We therefore implement a blocklist over both search results and shell commands.
The blocklist filters URLs and operations matching the target repository across GitHub, GitLab, raw-content endpoints, API endpoints, and direct git operations.
It also blocks solution-revealing artifacts associated with the target instance.
These safeguards force agents to use indirect external evidence, such as documentation, related projects, or general technical resources, rather than directly copying the gold solution.
With this setup, the experiments allow us to attribute performance changes to the availability and use of external information.

\section{Experiments}
\subsection{Experimental Setup}
We evaluate current code agents on \benchmark{} under four settings.
OpenHands~\citep{wang2025openhands} provides a standard open-source code-agent scaffold.
\agent{} is used as a controlled diagnostic baseline that keeps the same local coding workflow while adding web search and fetch tools.
We also evaluate the Codex~\citep{openai2025codexupgrades} harness (\texttt{v0.118.0}) in two prompt settings: its default coding prompt and a SearchSWE-style prompt that explicitly encourages search-aware coding.
The average score (AVG) is the mean of the four task-level scores, using Doc2Repo pass rate as its task score.
Table~\ref{tab:main_result} summarizes the main results.

\begin{table*}[t]
\centering
\small
\resizebox{0.95\textwidth}{!}{
\begin{tabular}{@{}lcccccc@{}}
\toprule
                                 & \textbf{CrossRepo} & \textbf{DomainFix} & \textbf{DepMigrate} & \multicolumn{2}{c}{\textbf{Doc2Repo}} &  \\ 
\cmidrule(lr){5-6}
\multirow{-2}{*}{\textbf{Model}} & \textbf{\%Resolved} & \textbf{\%Resolved} & \textbf{\%Resolved} & \textbf{Pass Rate} & \textbf{\#(Alm.) Corr.} & \multirow{-2}{*}{\textbf{AVG}} \\
\midrule

\multicolumn{7}{c}{\cellcolor{CornflowerBlue!15}\textit{\textbf{Codex}}} \\ 
\midrule
GPT-5.4 (xhigh) w/ SearchSWE Prompt 
& \textbf{55.17}\DeltaNotePlus{+8.9} 
& \textbf{61.11}\DeltaNotePlus{+19.4} 
& \textbf{48.59}\DeltaNotePlus{+4.2} 
& \textbf{61.74}\DeltaNotePlus{+0.1} 
& \textbf{(7) / 2} 
& \textbf{56.65}\DeltaNotePlus{+8.2} \\

GPT-5.4 (xhigh) w/ Default Prompt   
& \underline{46.23} 
& \underline{41.67} 
& \underline{44.38} 
& \underline{61.64} 
& \textbf{(7) / 2} 
& \underline{48.48} \\

\midrule
\multicolumn{7}{c}{\cellcolor{CornflowerBlue!15}\textit{\textbf{OpenHands}}} \\ 
\midrule

DeepSeek-V4-Pro(Max)~\citep{deepseekai2026deepseekv4}
& \underline{44.00} 
& \textbf{38.89} 
& \textbf{44.38} 
& \textbf{57.20} 
& (5) / 1
& \textbf{46.12} \\

GLM-5~\citep{zeng2026glm}
& \textbf{44.67} 
& 33.33 
& \underline{42.13} 
& \underline{56.76} 
& \underline{(7) / 3} 
& \underline{44.22} \\

Qwen3.5-Plus~\citep{qwen2026qwen35}
& 41.50 
& \textbf{38.89} 
& 41.01 
& 52.41 
& (3) / 1 
& 43.45 \\

Gemini 3 Pro~\citep{googledeepmind2026gemini3pro}          
& 41.50 
& 31.94 
& 41.81 
& 52.03 
& \textbf{(8) / 2} 
& 41.82 \\

GPT-5.4~\citep{openai2026gpt54}               
& 43.00 
& 27.78 
& 37.50 
& 56.30 
& (5) / 3 
& 41.15 \\

Kimi-K2.5~\citep{team2026kimi}             
& 40.50 
& \underline{34.72} 
& 39.89 
& 51.36 
& (3) / 1 
& 41.62 \\

Seed-Coder-2.0\citep{seed2025seed}            
& 39.50 
& 24.29 
& 33.15 
& 55.54 
& (5) / 2 
& 38.12 \\

MiniMax-M2.5~\citep{minimax2026m25}            
& 40.00 
& 25.00 
& 37.64 
& 46.57 
& (5) / 1 
& 37.30 \\

\midrule
\multicolumn{7}{c}{\cellcolor{CornflowerBlue!15}\textit{\textbf{SearchSWE}}} \\ 
\midrule

DeepSeek-V4-Pro(Max)~\citep{deepseekai2026deepseekv4}
& \textbf{48.50}\DeltaNotePlus{+4.5} 
& \textbf{43.06}\DeltaNotePlus{+4.2} 
& \textbf{47.19}\DeltaNotePlus{+2.8} 
& 56.16\DeltaNote{-1.0} 
& (5) / 2 
& \textbf{48.73}\DeltaNotePlus{+2.6} \\

GLM-5~\citep{zeng2026glm}       
& 43.88\DeltaNote{-0.8} 
& 35.94\DeltaNotePlus{+2.6} 
& \underline{47.13}\DeltaNotePlus{+5.0} 
& \textbf{60.05}\DeltaNotePlus{+3.3} 
& \underline{(7) / 3} 
& \underline{46.75}\DeltaNotePlus{+2.5} \\

Qwen3.5-Plus~\citep{qwen2026qwen35}
& 41.50 
& 34.72\DeltaNote{-4.2} 
& 39.89\DeltaNote{-1.1} 
& 54.90\DeltaNotePlus{+2.5} 
& (3) / 1 
& 42.75\DeltaNote{-0.7} \\

Gemini 3 Pro~\citep{googledeepmind2026gemini3pro}
& 41.12\DeltaNote{-0.4} 
& \underline{39.44}\DeltaNotePlus{+7.5} 
& 44.07\DeltaNotePlus{+2.3} 
& 50.73\DeltaNote{-1.3} 
& (4) / 2 
& 43.84\DeltaNotePlus{+2.0} \\

GPT-5.4~\citep{openai2026gpt54}                      
& \underline{45.00}\DeltaNotePlus{+2.0} 
& 31.94\DeltaNotePlus{+4.2} 
& 38.76\DeltaNotePlus{+1.3} 
& \underline{58.35}\DeltaNotePlus{+2.1} 
& \textbf{(7) / 4} 
& 43.51\DeltaNotePlus{+2.4} \\

Kimi-K2.5~\citep{team2026kimi}                    
& 43.00\DeltaNotePlus{+2.5} 
& 34.72 
& 37.64\DeltaNote{-2.3} 
& 52.22\DeltaNotePlus{+0.9} 
& (5) / 2 
& 41.90\DeltaNotePlus{+0.3} \\

Seed-Coder-2.0~\citep{seed2025seed}                   
& 43.15\DeltaNotePlus{+3.7} 
& 30.43\DeltaNotePlus{+6.1} 
& 35.80\DeltaNotePlus{+2.7} 
& 49.67\DeltaNote{-5.9} 
& (3) / 1 
& 39.76\DeltaNotePlus{+1.6} \\

MiniMax-M2.5~\citep{minimax2026m25}                 
& 39.00\DeltaNote{-1.0} 
& 31.94\DeltaNotePlus{+6.9} 
& 37.08\DeltaNote{-0.6} 
& 50.66\DeltaNotePlus{+4.1} 
& (3) / 0 
& 39.67\DeltaNotePlus{+2.4} \\

\bottomrule
\end{tabular}
}
\caption{
Main Results on \benchmark{}. 
We organize the results by three evaluation settings: Codex, OpenHands, and SearchSWE. 
For Codex, we report both the default prompt setting and the SearchSWE-style prompt setting to highlight the sensitivity of frontier search-code agents to prompt design.
\textcolor{darkred}{Red}/\textcolor{darkgreen}{green} values indicate gains/drops relative to the matched no-search setting: the default prompt for Codex and OpenHands for SearchSWE.
\textbf{Bold} and \underline{underlined} denote the best and second-best results across each evaluation settings.
}
\label{tab:main_result} 
\end{table*}

\subsection{Overall Performance on \benchmark{}}
\textbf{How close are current agents to solving \benchmark{}?}
The main result is that \benchmark{} remains far from saturated even for strong contemporary coding agents.
Under the OpenHands scaffold, the best model, DeepSeek-V4-Pro (Max), reaches only 46.12 average score.
The Codex harness with GPT-5.4 (xhigh) performs better, reaching 48.48 with the default prompt, but this still leaves substantial headroom across all four task families.
When prompted with the SearchSWE-style workflow, the same Codex configuration reaches the strongest overall result, 56.65, yet the benchmark is still not close to being solved.

The task-level pattern shows that \benchmark{} is not difficult for a single reason.
CrossRepo stresses external software knowledge, DomainFix requires domain-specific reasoning, and DepMigrate demands coordinated repository-level edits.
Doc2Repo exposes a different limitation: its pass rate can overstate complete success, since even the best configuration produces only 2 fully correct repositories out of 50.
Overall, current agents often make meaningful partial progress, but still fail to deliver reliable end-to-end solutions across broader knowledge and resolution scopes.

\begin{figure}[t]
    \centering
    \includegraphics[width=0.7\linewidth]{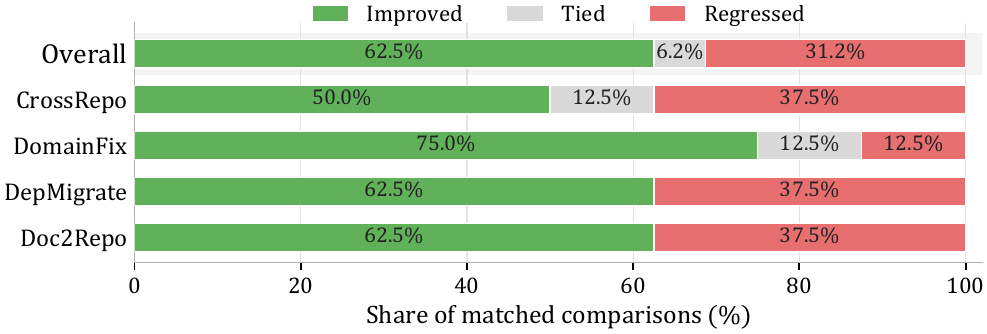}
    \caption{
    Task-wise decomposition of SearchSWE outcomes relative to OpenHands.
    Each task row aggregates the eight matched model pairs. The Overall row aggregates all 32 task-model comparisons.
    }
    \label{fig:search_gain_decomposition}
\end{figure}

\subsection{Search Helps, but Integration Matters}
\textbf{Does external search access improve coding performance?}
The answer is broadly yes, but the gains are limited and harness-dependent.
Compared with OpenHands, \agent{} improves seven of eight evaluated models, with the strongest \agent{} configuration reaching 48.73.
At the task level, 20 of 32 paired comparisons improve, while 31.2\% regress, showing that search access is useful but not uniformly reliable.
Figure~\ref{fig:search_gain_decomposition} decomposes these outcomes by task.
Overall, DomainFix benefits most, consistent with its reliance on externally retrievable domain knowledge, whereas CrossRepo, DepMigrate, and Doc2Repo retain substantial regressions.

\begin{figure}[t]
    \centering
    \includegraphics[width=0.7\linewidth]{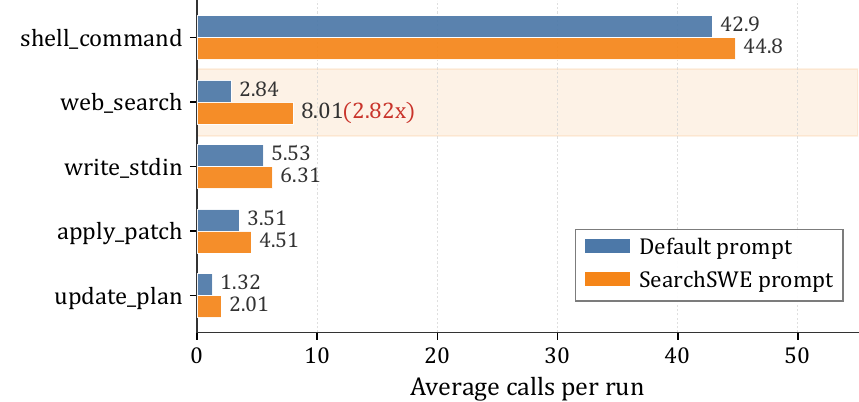}
    \caption{
    Average tool calls per instance for Codex with GPT-5.4 (xhigh) under the default prompt and the SearchSWE-style prompt.
    The search-oriented prompt substantially increases web-search use while keeping local coding-tool usage comparable.
    }
    \label{fig:codex_prompt_tool_usage}
\end{figure}

The Codex comparison complements this aggregate view.
Using the same GPT-5.4 (xhigh) model and harness, the SearchSWE-style prompt improves AVG from 48.48 to 56.65, with gains concentrated on DomainFix (+19.4) and CrossRepo (+8.9).
Figure~\ref{fig:codex_prompt_tool_usage} shows that the prompt increases average web-search calls by about 2.8$\times$ while preserving a similar local coding-tool profile.
This suggests that search access is useful but not self-activating: frontier harnesses may still need explicit guidance to turn external information into local, version-compatible code changes.

\begin{figure*}[t]
    \centering
    \begin{subfigure}[t]{0.49\textwidth}
        \centering
        \includegraphics[width=\linewidth]{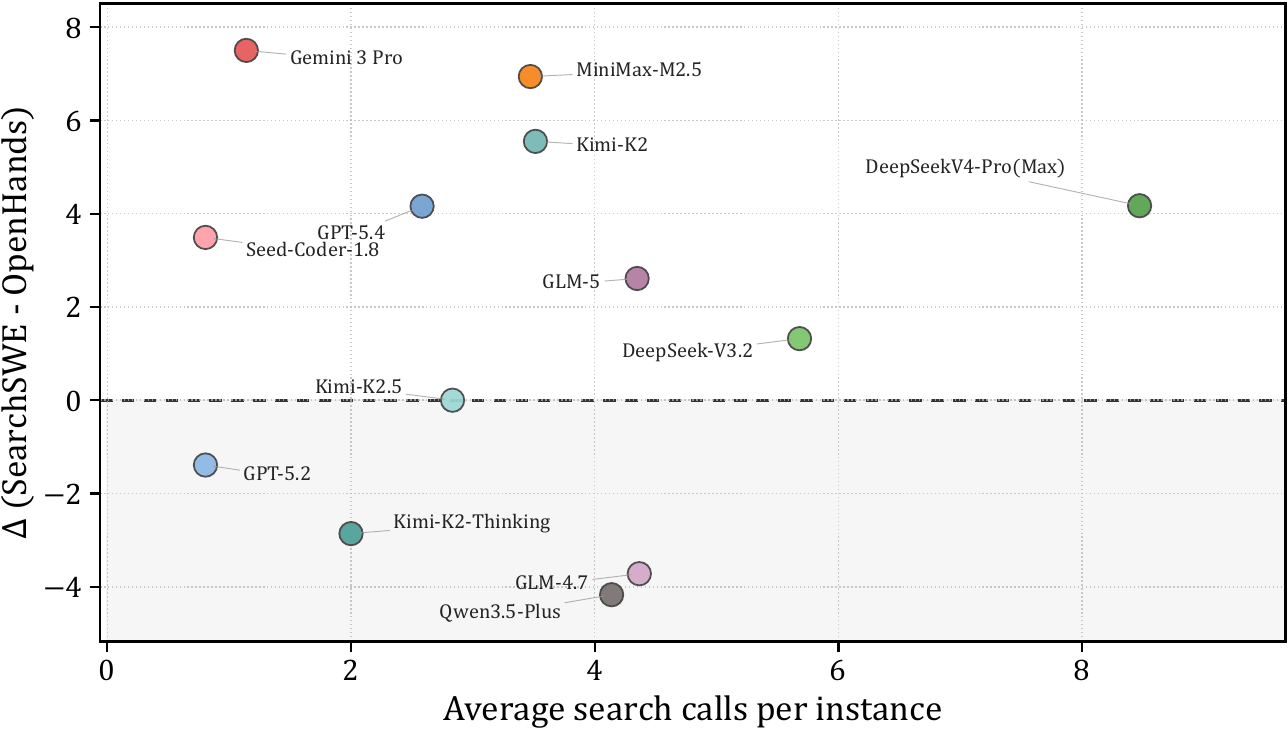}
        \caption{DomainFix: largest average gain.}
        \label{fig:search_effort_domainfix}
    \end{subfigure}
    \hfill
    \begin{subfigure}[t]{0.49\textwidth}
        \centering
        \includegraphics[width=\linewidth]{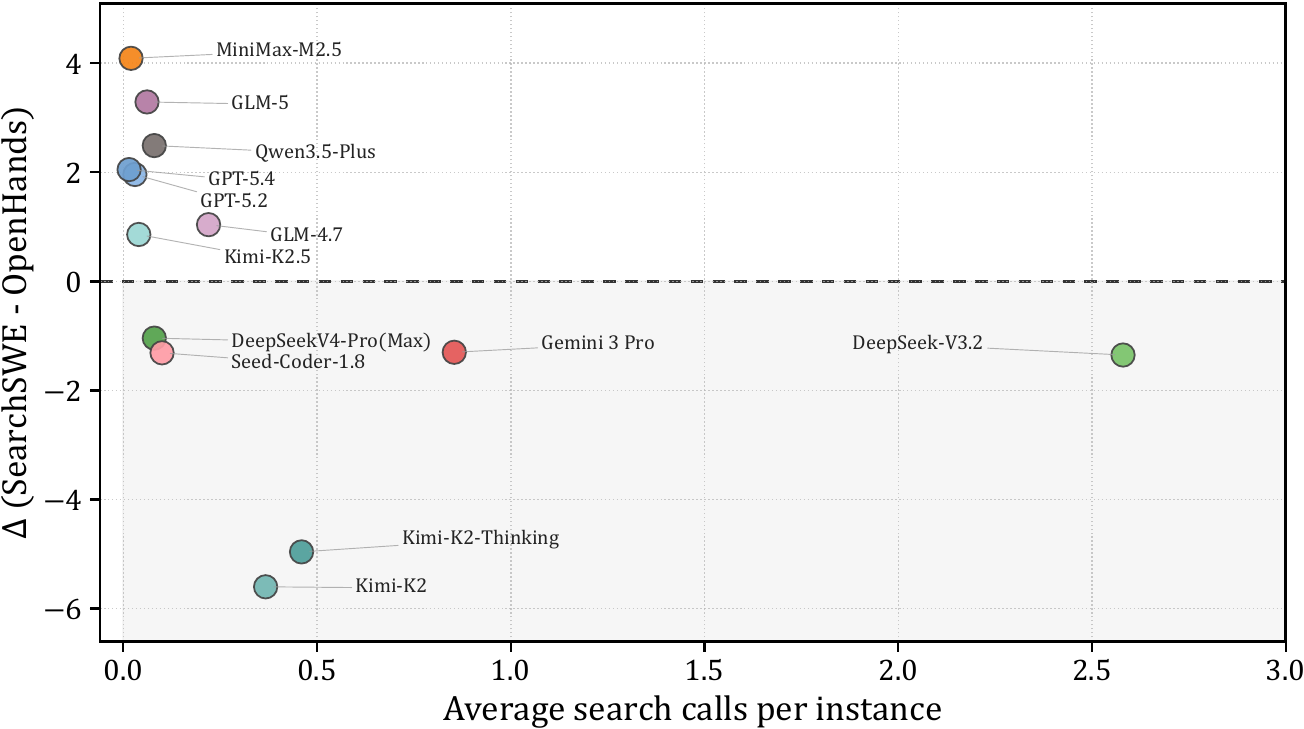}
        \caption{Doc2Repo: weakest average gain.}
        \label{fig:search_effort_doc2repo}
    \end{subfigure}
    \caption{
    Search effort versus \agent{} gain on two contrasting tasks.
    Each point is an evaluated model with matched OpenHands and \agent{} runs.
    The y-axis shows task-score of \agent{} delta over OpenHands.
    Within each task, more search calls do not monotonically imply larger gains.
    }
    \label{fig:search_effort_gain}
\end{figure*}

\textbf{Is more search necessarily better?}
Figure~\ref{fig:search_effort_gain} compares search-related calls with \agent{} gains on DomainFix and Doc2Repo, the tasks with the strongest and weakest average search gains.
DomainFix contains many positive-gain points, while Doc2Repo remains clustered around small or negative gains.
Within each task, more search calls do not monotonically imply larger gains.
This suggests that the bottleneck is not how often agents search, but whether they retrieve relevant evidence and ground it in the local task context.

\begin{figure*}[t]
    \centering
    \begin{subfigure}[t]{0.49\textwidth}
        \centering
        \includegraphics[width=\linewidth]{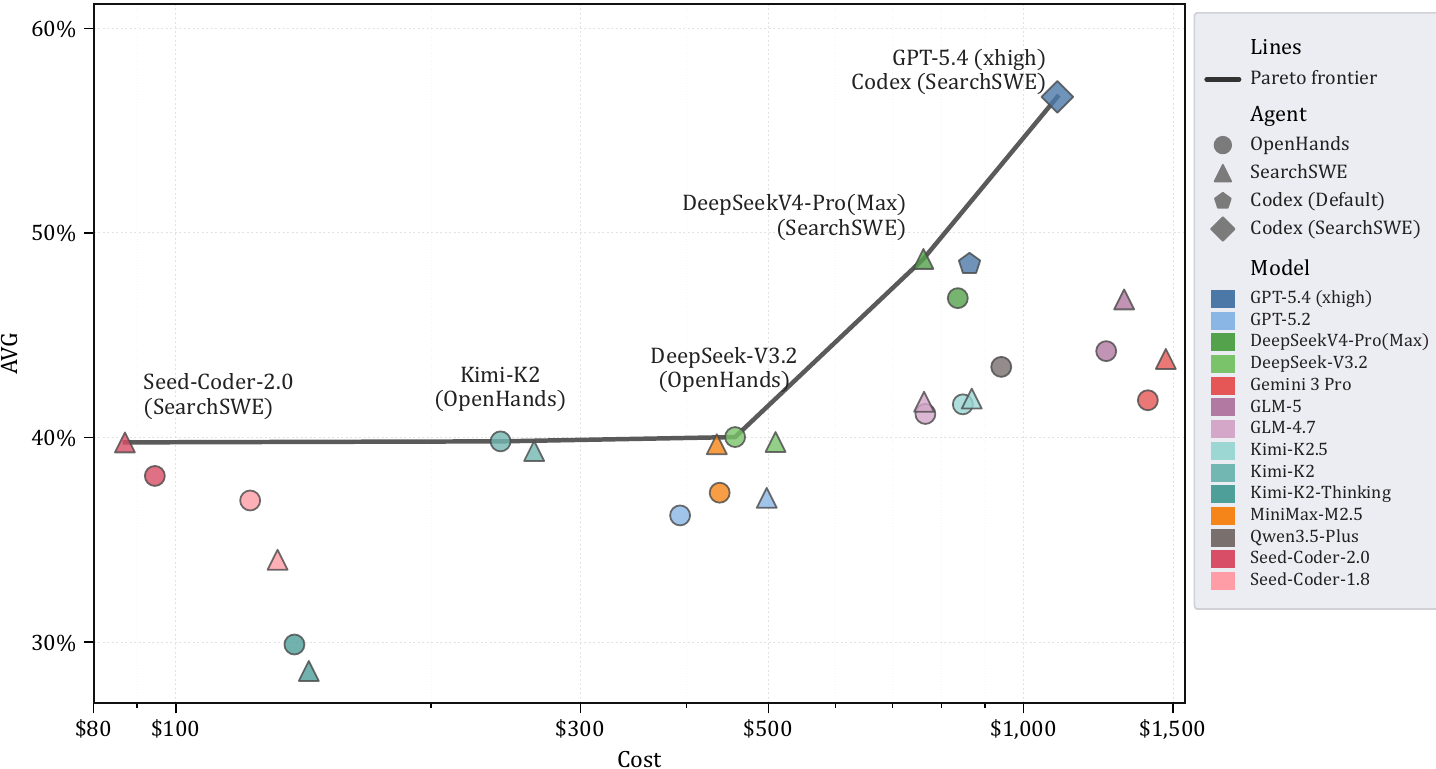}
        \caption{All Cost (\$) vs. Performance.}
        \label{fig:cost_performance}
    \end{subfigure}
    \begin{subfigure}[t]{0.49\textwidth}
        \centering
        \includegraphics[width=\linewidth]{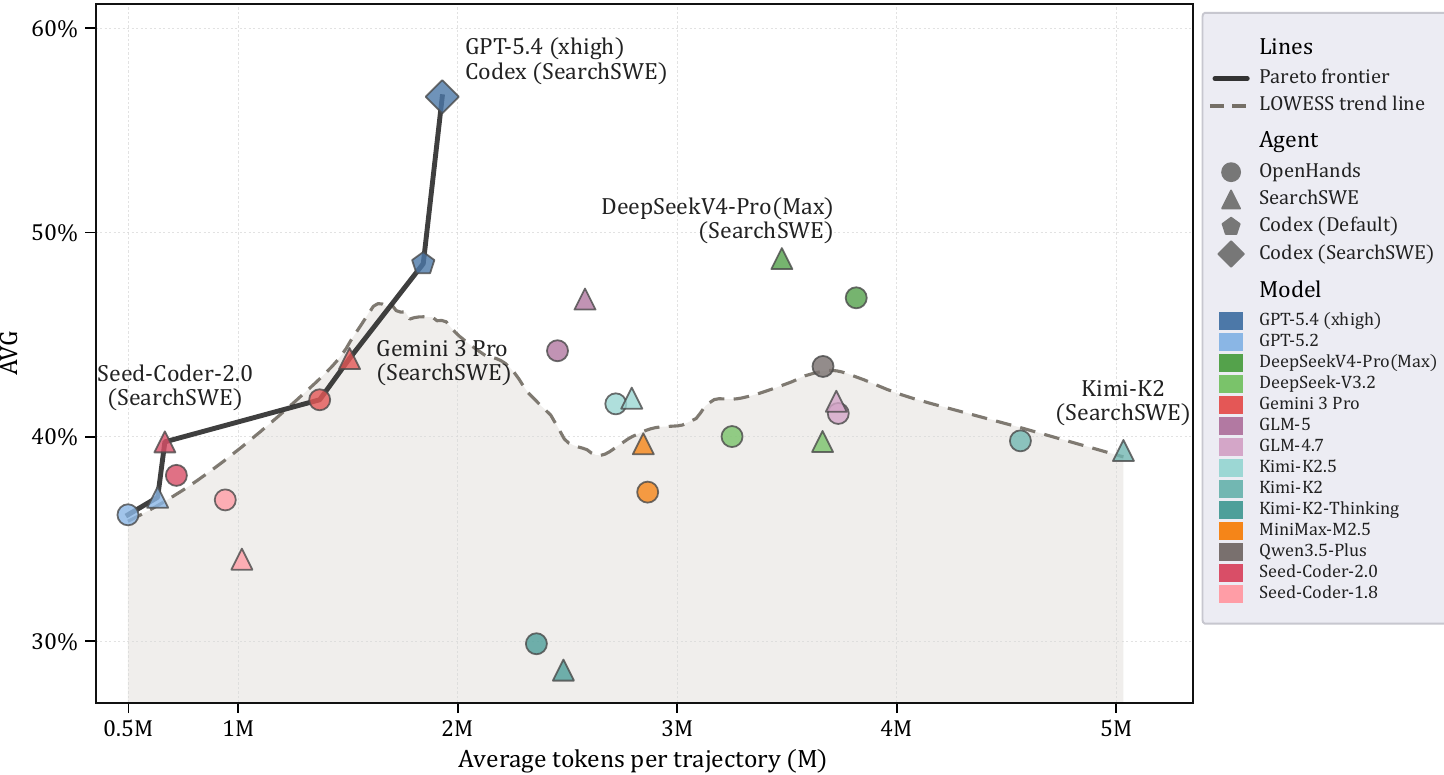}
        \caption{Average tokens (M) vs. Performance.}
        \label{fig:token_performance}
    \end{subfigure}
    \caption{
    Cost and token-budget efficiency on \benchmark{}.
    Solid black lines denote the Pareto frontier among evaluated configurations.
    In the token plot, the dashed LOWESS curve summarizes the local performance trend: points below it are less token-productive than configurations with similar token budgets, and the high-token flattening shows that longer trajectories alone do not reliably improve performance.
    }
    \label{fig:budget_efficiency}
\end{figure*}

\textbf{Why search remains insufficient.}
The qualitative cases in Appendix~\ref{app:case_study} identify three failure sites in the search-to-code pipeline: evidence retrieved at the wrong granularity, external knowledge not grounded in the local dependency version, and keyword-matched but semantically unrelated results contaminating the coding context.
Thus, the limitation is not access to search alone, but the ability to decide which external evidence is trustworthy and locally actionable.
Robust search-augmented coding still requires source discrimination, version grounding, and local verification before retrieved evidence becomes a code change.

\subsection{Cost and Token Budget Are Not Enough}
\textbf{Are stronger results simply a consequence of spending more money or tokens?}
Figure~\ref{fig:budget_efficiency} suggests not.
Higher API cost does not reliably correspond to higher performance: some lower-cost configurations, such as Seed-Coder-2.0~\citep{seed2025seed} under \agent{}, remain competitive, while several high-cost configurations do not lie on the performance frontier.
This indicates that benchmark performance is not merely purchased through larger inference budgets.

The token-performance view shows a similar pattern from trajectory length.
The Pareto frontier highlights the token-efficient configurations: many higher-token runs fall below this frontier without corresponding performance gains.
The LOWESS trend rises in the low-token region but flattens or declines at higher token budgets, indicating that longer trajectories do not reliably translate into better task resolution.
Many high-token runs appear to spend budget on repeated exploration, noisy search, or ineffective repair loops rather than successful fixes.
Taken together, these results point to token productivity rather than token volume as the key bottleneck.
More budget can help when paired with a strong model, scaffold, and search-aware workflow, but current agents do not automatically convert additional tokens into grounded search, correct local reasoning, and targeted edits.



\section{Conclusion}
\label{sec:conclusion}
We introduced \benchmark{}, a 500-instance benchmark for evaluating code agents beyond single-repository bug fixing, spanning cross-repository reasoning, domain-specific repair, dependency migration, and repository construction from specifications.
Our evaluation shows that stronger models and harnesses improve performance, but current agents remain far from saturating these settings.
Using \agent{} as a controlled diagnostic baseline, we find that external search is genuinely useful, especially when missing knowledge is externally retrievable, but its gains are uneven and depend on whether agents can ground retrieved evidence in the local repository, dependency version, and task specification.
These results suggest that deep search for coding remains an open problem: progress requires agents that not only retrieve information, but also decide when to trust it, when to ignore it, and how to convert it into precise, executable code changes.

\bibliography{main}
\appendix
\newpage

\addcontentsline{toc}{section}{Appendix}

\begin{center}
{\Large\bfseries\color{AweAIblue} Contents of Appendix}
\end{center}

\startcontents[appendix]
\printcontents[appendix]{}{1}{\setcounter{tocdepth}{2}}

\vspace{2em}

\clearpage

\section{Additional Related Work}
\label{app:additional_related_work}
\textbf{Deep research agents.}
Recent deep research systems extend LLM agents with iterative web search, browsing, source synthesis, and long-horizon information gathering~\citep{dr,google_dr,team2025tongyi,qiao2025webresearcher,chen2025cpo,Perplexity,iterresearch}.
These systems demonstrate strong progress in open-domain research workflows, but their objectives are usually information synthesis rather than producing executable code changes under repository-local constraints.
\benchmark{} uses search-augmented coding to study a different setting: retrieved information must be converted into precise, version-compatible, and test-passing code edits.

\textbf{Code agents and agent training.}
A parallel line of work improves code agents through stronger scaffolds, agent-computer interfaces, trajectory data, supervised fine-tuning, and reinforcement learning~\citep{yang2024sweagent,moatless2024,wang2025openhands,xia2024agentless,yang2025swe,guo2025swefactory,SWESwiss2025,tao2026swe,zhao2026immersion}.
Most of this work optimizes agents for repository-local issue resolution, especially SWE-bench-style tasks~\citep{wei2025swe,yang2025kimi,yang2025scaling,ma2025swe,jain2025r2e,wang2025swe}.
Our analysis complements these efforts by evaluating whether agents can combine local coding with external information seeking across tasks that require broader knowledge or broader resolution scope.

\section{Qualitative Analysis of Search-Augmented Coding Failures}
\label{app:case_study}

This section examines why external search does not always translate into better coding performance.
The three cases below are not isolated anecdotes; they illustrate recurring breaks in the pipeline from external evidence to local code changes.
Table~\ref{tab:case_study_summary} summarizes the failure modes.

At a high level, search-augmented coding fails in three places.
First, \textit{the relevant evidence may not be what search engines rank highly}: agents may receive user-facing documentation when they need source-level implementation logic.
Second, \textit{retrieved information must be grounded in the local repository state}: a correct pattern for a newer library version can be wrong for the pinned environment.
Third, \textit{agents must filter semantically irrelevant results}: keyword overlap can pull in authoritative but unrelated sources that contaminate the coding context.
Together, these cases explain why search access is useful but insufficient on \benchmark{}.

\begin{table*}[t]
\centering
\small
\resizebox{\textwidth}{!}{
\begin{tabular}{@{}p{2.8cm}p{3.0cm}p{3.8cm}p{3.6cm}p{2.6cm}@{}}
\toprule
\textbf{Failure Mode} & \textbf{Instance} & \textbf{Search Behavior} & \textbf{Local Failure} & \textbf{Takeaway} \\
\midrule
Evidence availability &
unidata\_siphon\_pr234 &
Search returns high-level documentation instead of source-level backend logic &
Agent implements a brittle interpretation of an ambiguous API description &
Search must recover evidence at the right technical granularity \\
\midrule
Version grounding &
behave\_behave-django\_pr162 &
Agent searches around an unsupported newer Django assumption instead of local constraints &
Agent applies a method pattern incompatible with the repository's legacy test lifecycle &
External evidence must be filtered through local dependency versions \\
\midrule
Evidence filtering &
abravalheri\_validate-pyproject\_pr105 &
Keyword-matched results drift into unrelated domains due to overloaded terminology &
Agent falls back to a generic plugin-registration pattern that creates test side effects &
Agents need semantic source discrimination, not only retrieval \\
\bottomrule
\end{tabular}
}
\caption{Summary of search-augmented coding failure modes analyzed in the case studies.}
\label{tab:case_study_summary}
\end{table*}

\subsection{Failure Mode I: Source-Level Evidence Is Hard to Retrieve}
\label{app:case_study_information_landscape}

\textbf{Failure mode.}
Some coding tasks require low-level artifacts such as source files, commit diffs, or backend implementation logic, but web search often prioritizes curated, user-facing documentation.
The result may be conceptually relevant but too imprecise to support a robust code change.

\textbf{Case.}
In \texttt{unidata\_siphon\_pr234}, the agent needs to extend \texttt{IAStateUpperAir} so it can fetch data for all stations at once.
The behavior is enabled by an update in the Iowa Environmental Mesonet backend, but the target repository does not contain the backend implementation.
To implement the change correctly, the agent needs to understand the exact parameter handling expected by the backend CGI script.

\textbf{What went wrong.}
The agent correctly recognizes that local context is insufficient and searches for the backend artifact using a query targeting \texttt{akrherz/iem} and \texttt{raob.py}.
However, as shown in Table~\ref{tab:info_mismatch}, search returns a high-level API help page rather than the backend source code.
The help page says the service can be queried with ``just a timestamp,'' which is directionally useful but syntactically ambiguous: it does not specify whether to remove the \texttt{station} parameter, set it to \texttt{None}, pass a wildcard, or follow additional backend-specific error-handling logic.

\begin{table*}[h]
\centering
\begin{small}
\begin{tabular}{@{}p{3cm} p{5.5cm} p{5.5cm}@{}}
\toprule
\textbf{Component} & \textbf{Agent's Intent / Target} & \textbf{Search Engine Response} \\
\midrule
\textbf{Target Artifact} & \textbf{Backend Source Code} & \textbf{User-Facing Documentation} \\
& (\texttt{akrherz/iem/.../raob.py}) & (\texttt{.../json/raob.py?help}) \\
\addlinespace
\textbf{Information Type} & \textbf{Precise Logic} & \textbf{Ambiguous Natural Language} \\
& Explicit conditionals, error handling, and parameter parsing logic. & High-level description: ``approach this service... with \textcolor{red}{\textbf{just a timestamp}}.'' \\
\addlinespace
\textbf{Resulting Action} & \textit{(Hypothetical)} Implement robust parameter negotiation mirroring backend logic. & \textit{(Actual)} Implemented a brittle solution based on literal interpretation of "just a timestamp". \\
\bottomrule
\end{tabular}
\end{small}
\vspace{-0.5em}
\caption{Analysis of the Information Landscape Mismatch in \texttt{unidata\_siphon\_pr234}. The search engine's bias towards curated content obscures the precise logic required for robust code implementation.}
\label{tab:info_mismatch}
\end{table*}

The agent then implements the most literal interpretation of the documentation by omitting the station parameter in selected cases.
This solves part of the intended behavior, but it misses edge-case handling encoded in the source-level backend logic.
The resulting patch fails comprehensive tests such as \texttt{test\_no\_future\_data\_with\_pressure\_iastate}.

\begin{lessonbox}
The failure is not simply that search failed to find anything relevant.
Rather, search found evidence at the wrong granularity, and the agent did not recognize that the retrieved documentation was insufficient for implementation.
Search-augmented coding therefore requires agents to judge whether retrieved evidence is precise enough for source-level changes.
\end{lessonbox}

\subsection{Failure Mode II: Retrieved Knowledge Must Be Version-Grounded}
\label{app:case_study_version_consistency}

\textbf{Failure mode.}
External information can conflict with the version constraints of the local repository.
For maintenance tasks, the right answer is not necessarily the newest API pattern, but the pattern compatible with the installed dependency versions and the repository's existing architecture.

\textbf{Case.}
In \texttt{behave\_behave-django\_pr162}, the agent must fix a fixture-loading issue in a repository pinned to legacy Django versions.
The relevant constraints are available locally through project configuration and installed packages.
The task therefore requires grounding any external information in the actual environment before editing code.

\textbf{What went wrong.}
Instead of first verifying the installed Django version, the agent forms an unsupported assumption about a newer Django setting and searches for evidence around that assumption.
As Table~\ref{tab:version_conflict} shows, this pushes the agent toward a modern class-method pattern for lifecycle hooks.
That pattern conflicts with the local repository, where \texttt{\_pre\_setup} participates in a legacy instance-method lifecycle.

\begin{table*}[h]
    \centering
    \begin{small}
    \begin{tabular}{@{}p{3.5cm} p{6.5cm} c@{}}
        \toprule
        \textbf{Context Source} & \textbf{Code Pattern \& Logic} & \textbf{Status} \\
        \midrule
        \textbf{Local Environment} \newline (Django 2.2/3.x) & \textbf{Instance Method (Legacy)} \newline \texttt{def \_pre\_setup(self): ...} \newline \textit{Constraint: Must maintain state on the test instance.} & \textbf{Ignored} \\
        \midrule
        \textbf{Agent Search} \newline (Hallucinated "v5.2") & \textbf{Class Method (Modern/Future)} \newline \texttt{@classmethod} \newline \texttt{def \_pre\_setup(cls): ...} \newline \textit{Bias: Assumes newer patterns apply universally.} & \textbf{Prioritized} \\
        \midrule
        \textbf{Implementation} \newline (The Error) & \textbf{Signature Mismatch} \newline Agent applies \texttt{@classmethod} to legacy hook. \newline \textit{Result: Breaks MRO and instance state access.} & \xmark \\
        \bottomrule
    \end{tabular}
    \end{small}
    \vspace{-0.5em}
    \caption{Analysis of Version Conflict in \texttt{behave\_behave-django\_pr162}. The agent prioritizes hallucinated future specifications over explicit local constraints.}
    \label{tab:version_conflict}
\end{table*}

The implementation then changes \texttt{\_pre\_setup} into a \texttt{@classmethod}.
This breaks the expected interaction with instance-level state and the inherited test lifecycle, causing the test suite to fail.
The key error is not merely a bad search query; it is the absence of a local-version check before applying retrieved or recalled patterns.

\begin{lessonbox}
Search for coding is not a search for the most recent answer.
It is a search for evidence compatible with the local environment.
Agents need to treat local dependency versions, installed APIs, and inherited code structure as constraints that filter external knowledge.
\end{lessonbox}

\subsection{Failure Mode III: Keyword Matches Can Contaminate Context}
\label{app:case_study_noise_search}

\textbf{Failure mode.}
Technical terms are often overloaded across domains.
For niche libraries, a reasonable query can retrieve high-ranking results that match the keywords but not the repository's semantic context.
If an agent fails to reject these results, search becomes a source of context contamination.

\textbf{Case.}
In abravalheri\_validate-pyproject\_pr105, the agent must implement support for \texttt{repo-review}, where checks are grouped into ``families.''
The agent searches for \texttt{"repo-review define checks fixtures families"}, a query that is reasonable given the task wording.

\textbf{What went wrong.}
As shown in Table~\ref{tab:search_noise}, the first result is relevant, but later results drift into unrelated domains such as Autodesk Revit and RelativityOne.
The drift is caused by overloaded terms such as ``family'' and ``review,'' combined with the low web footprint of the target library.
Rather than isolating the single relevant source and discarding unrelated entries, the agent fails to recover a precise integration pattern.

\begin{table*}[h]
    \centering
    \begin{small}
    \begin{tabular}{@{}p{3.5cm} p{1.8cm} p{7.5cm} c@{}}
        \toprule
        \textbf{Search Result Title} & \textbf{Domain} & \textbf{Snippet Content} & \textbf{Rel.} \\
        \midrule
        Families - repo-review 0.12.4.dev15 ... & \textbf{Target} & ``Families are a set of simple strings that group together similar checks...'' & \cmark \\
        \midrule
        Writing Family Instances with the Revit Writer & Architecture (BIM) & ``...highlight some notable points on \textbf{Revit family instances} and the generic API...'' & \xmark \\
        \midrule
        RelativityOne - User Guide & Legal Tech & ``...launch the Review interface from the Documents and \textbf{Family card}...'' & \xmark \\
        \midrule
        Maryland Workforce Innovation... & Gov. Policy & ``...conducts an onsite \textbf{review} to examine all resource documents...'' & \xmark \\
        \bottomrule
    \end{tabular}
    \end{small}
    \vspace{-0.5em}
    \caption{Analysis of Retrieval Noise in \texttt{abravalheri\_validate-pyproject\_pr105}. The polysemy of the term "Family" causes the search engine to drift into unrelated technical domains.}
    \label{tab:search_noise}
\end{table*}

The agent then falls back to a generic Python plugin-registration prior using \texttt{entry\_points}.
This change is plausible in isolation but incompatible with the target tests, which already register plugins through fixtures.
The implementation registers the plugin twice and triggers a side-effect failure:

\begin{lstlisting}[basicstyle=\small\ttfamily, frame=tb, breaklines=true, numbers=none]
________ TestDisable.test_parse ________
> assert len(params.plugins) == 1
E AssertionError: assert 2 == 1
E  +  where 2 = len([<PluginWrapper...>, <PluginWrapper...>])
\end{lstlisting}

\begin{lessonbox}
The challenge is not only retrieving information, but discriminating which sources belong to the repository's semantic context.
Search-augmented coding requires agents to reject high-ranking but out-of-domain evidence and to verify that an externally suggested pattern is locally actionable before modifying code.
\end{lessonbox}
\section{More Implementation Details}

\subsection{Data and Code Availability}
We submit anonymized supplementary material containing the benchmark data, evaluation code, and documentation needed to inspect \benchmark{} and reproduce the evaluation protocol.
Public release links are omitted during anonymous review and will be provided after the review process.

\subsection{Agent and Tool Configuration}
\begin{table}[t]
    \centering
        \small
        \begin{tabular}{ll}
            \toprule
            \textbf{Research Field} & \textbf{Repositories} \\
            \midrule
            \rowcolor{CornflowerBlue!15} \multicolumn{2}{c}{\textbf{Scientific Computing}} \\
            Astronomy               & astroplan \\
            Bioinformatics          & biotite, Biopython \\
            Computational chemistry & cclib \\
            Plasma physics          & PlasmaPy \\
            Quantum physics         & qutip \\
            Seismology              & obspy \\
            \midrule
            \rowcolor{CornflowerBlue!15} \multicolumn{2}{c}{\textbf{Engineering}} \\
            Convex optimization     & cvxpy \\
            Geospatial              & geopandas \\
            Materials science       & pymatgen \\
            Molecular dynamics      & mdanalysis \\
            Photonic IC design      & gdsfactory \\
            \bottomrule
        \end{tabular}
        \vspace{-0.5em}
        \captionof{table}{Research fields and repositories included in the DomainFix task.}
        \label{tab:domain_repos}
\end{table}
\textbf{Evaluation Details.}
For both OpenHands and \agent{}, we set the maximum number of interaction turns to 200. The maximum context length is determined by each model's native limit.
OpenHands~\citep{wang2025openhands} is equipped with three core tools: \texttt{ExecuteBashTool} for command execution, \texttt{StrReplaceEditorTool} for file editing, and \texttt{FinishTool} for task completion.
\agent{} extends this toolset with two additional capabilities: \texttt{SearchTool} for web search and \texttt{BrowserTool} for webpage browsing.
The search functionality is powered by Google Search via SerpAPI\footnote{\url{https://serpapi.com/}}, while the browser tool utilizes Jina Reader\footnote{\url{https://jina.ai/}} for content extraction, with DeepSeek-V3.2 serving as the summarization model.
For Codex experiments, we use Codex (\texttt{v0.118.0}).

\textbf{Controlled comparison.}
The comparison between OpenHands and \agent{} is designed to isolate the effect of external information access.
Both settings use the same local execution environment, editing interface, shell access, test commands, and maximum interaction budget.
\agent{} differs only by exposing web search and fetch tools and by using prompts that instruct the agent how to incorporate retrieved evidence into coding and verification.

\subsection{\benchmark{} Dataset Details}
\textbf{Repository Details of DomainFix.}\label{app:domainfix_repos}
Table~\ref{tab:domain_repos} lists the 11 research fields and corresponding repositories included in the DomainFix task.
These repositories span diverse scientific domains, from astronomy and quantum physics to bioinformatics and materials science, each requiring specialized domain knowledge to resolve issues.

\begin{figure}
    \centering
    \includegraphics[width=0.5\linewidth]{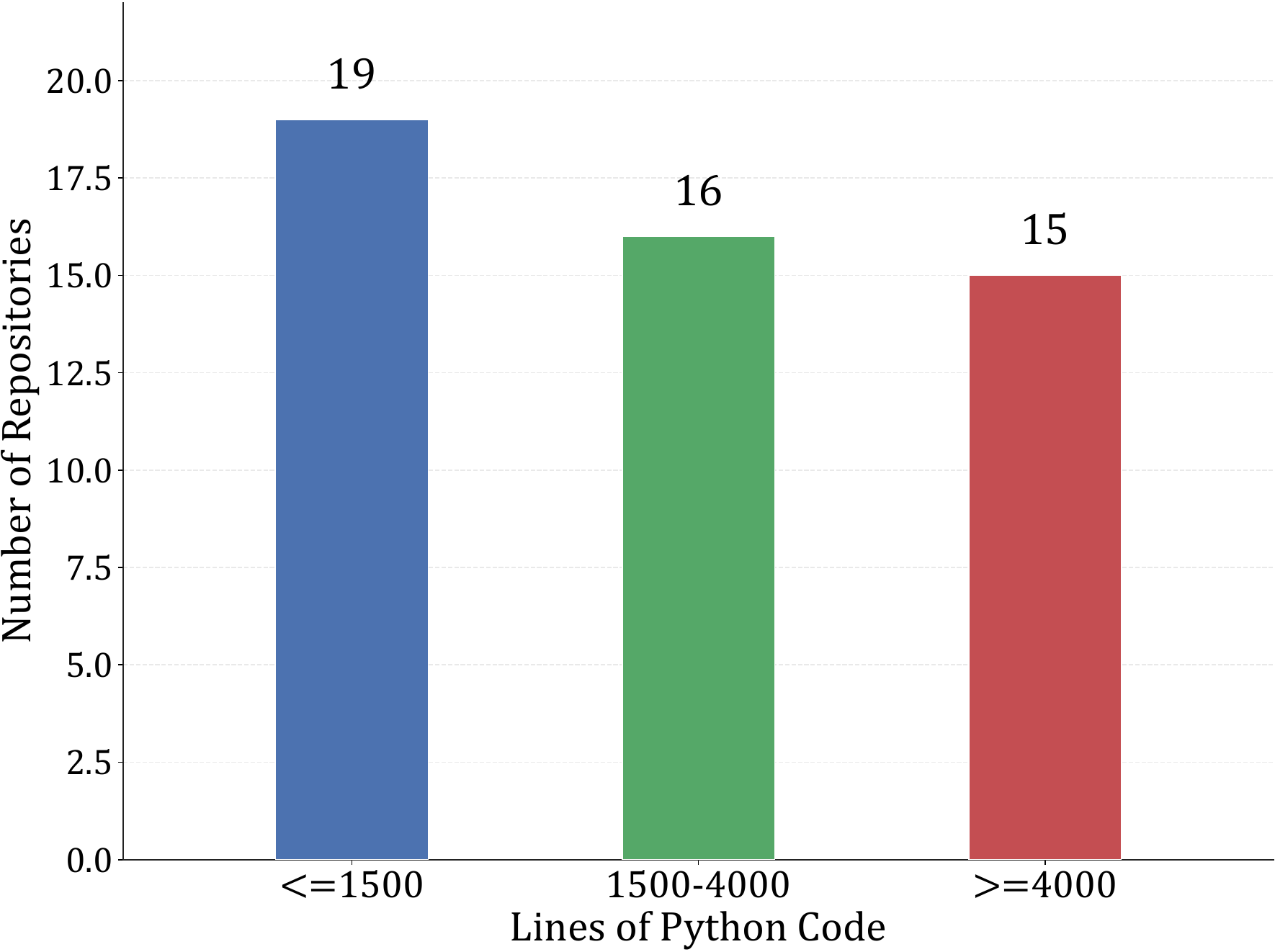}
    \vspace{-0.5em}
    \caption{Distribution of lines of code across the 50 Doc2Repo repositories.}
    \label{fig:doc2repo_dist}
\end{figure}

\begin{figure*}[h]
    \centering
    \includegraphics[width=\linewidth]{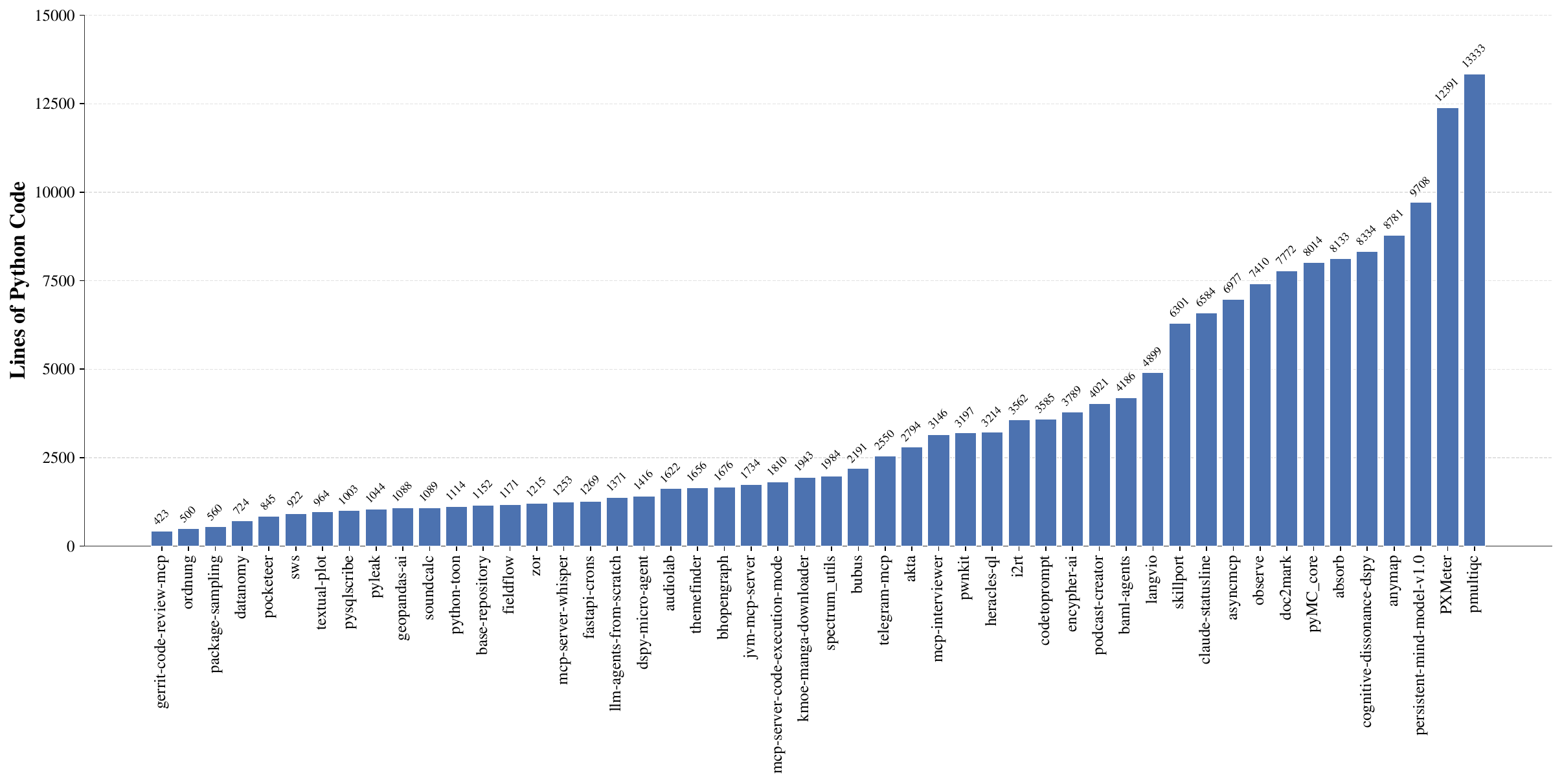}
    \caption{Lines of code for each Doc2Repo repository, sorted by size.}
    \label{fig:doc2repo_lines}
\end{figure*}

\textbf{Doc2Repo Repository Statistics.}\label{app:doc2repo_python_code_lines}
Figure~\ref{fig:doc2repo_dist} and Figure~\ref{fig:doc2repo_lines} present the code size statistics for the 50 repository instances in the Doc2Repo task.
The repositories range from approximately 400 to over 13,000 lines of code, with the majority (31 out of 50) exceeding 1,500 lines.
This distribution demonstrates that Doc2Repo poses a substantial challenge, requiring agents to generate non-trivial, real-world scale codebases rather than simple toy examples.

\textbf{Data Format.}\label{app:data_format}
Each instance in \benchmark{} is stored as a JSON object containing the following fields (Table~\ref{tab:data_fields}).
For Doc2Repo, the \texttt{problem\_statement} field contains the generated repository specification, and evaluation uses the complete adapted test suite rather than a P2P/F2P split.
\begin{table*}[h]
\centering
\small
\begin{tabular}{lp{10cm}}
\toprule
\textbf{Field} & \textbf{Description} \\
\midrule
\texttt{instance\_id} & Unique identifier for the instance, typically in the format \texttt{user\_repo\_prN}. \\
\texttt{dataset\_id} & Dataset identifier (e.g., \texttt{realswe\_bench}). \\
\texttt{task} & Task category: \texttt{crossrepo}, \texttt{domainfix}, \texttt{depmigrate}, or \texttt{doc2repo}. \\
\texttt{user} & GitHub organization or user name. \\
\texttt{repo} & Repository name. \\
\texttt{language} & Primary programming language of the repository. \\
\texttt{workdir} & Working directory path inside the Docker container. \\
\texttt{image\_url} & Docker image URL for the evaluation environment. \\
\texttt{patch} & Gold patch (ground truth solution) in unified diff format. \\
\texttt{pr\_commit} & Commit hash of the pull request that resolved the issue. \\
\texttt{parent\_commit} & Commit hash of the codebase before the fix (evaluation starting point). \\
\texttt{problem\_statement} & Natural language description of the issue to be resolved. \\
\texttt{github\_url} & URL to the original GitHub repository. \\
\texttt{pre\_commands} & Shell commands to initialize the environment before evaluation. \\
\texttt{FAIL\_TO\_PASS} & List of test cases that should change from failing to passing after the fix. \\
\texttt{PASS\_TO\_PASS} & List of test cases that should remain passing after the fix. \\
\bottomrule
\end{tabular}
\caption{Data fields for each instance in \benchmark{}.}
\label{tab:data_fields}
\end{table*}

\subsection{Additional Evaluation Results}
\label{app:additional_results}

Table~\ref{tab:additional_results} reports additional OpenHands and \agent{} results for models not included in the main table.
These runs support the same qualitative pattern as the main experiments: search access often improves performance, but the gains remain task- and model-dependent, with regressions on some task-model pairs.
\begin{table*}[t]
\centering
\scriptsize
\resizebox{0.95\textwidth}{!}{
\begin{tabular}{@{}lcccccc@{}}
\toprule
                                 & \textbf{CrossRepo} & \textbf{DomainFix} & \textbf{DepMigrate} & \multicolumn{2}{c}{\textbf{Doc2Repo}} &  \\ 
\cmidrule(lr){5-6}
\multirow{-2}{*}{\textbf{Model}} & \textbf{\%Resolved} & \textbf{\%Resolved} & \textbf{\%Resolved} & \textbf{Pass Rate} & \textbf{\#(Alm.) Corr.} & \multirow{-2}{*}{\textbf{AVG}} \\
\midrule

\multicolumn{7}{c}{\cellcolor{CornflowerBlue!15}\textit{\textbf{OpenHands}}} \\ 
\midrule
GLM-4.7~\citep{zai2025glm47}
& \underline{40.20}
& \textbf{36.11}
& \textbf{39.89}
& 48.40
& \underline{(3) / 1}
& \textbf{41.15} \\

DeepSeek-V3.2~\citep{liu2025deepseek}
& 38.00
& \underline{30.56}
& 36.52
& \textbf{54.99}
& (3) / 0
& \underline{40.02} \\

Kimi-K2~\citep{team2025kimi}
& 37.00
& 27.78
& \underline{39.53}
& \underline{54.91}
& \textbf{(6) / 2}
& 39.81 \\

GPT-5.2~\citep{openai2025gpt52}
& 33.00
& 23.61
& 34.27
& 53.89
& \textbf{(6) / 2}
& 36.19 \\

Seed-Coder-1.8~\citep{seed2025seed}
& \textbf{41.92}
& 18.57
& 31.46
& 42.71
& (1) / 0
& 33.67 \\

Qwen3-Coder-Plus~\citep{qwen3-coder-2025}
& 19.19
& 5.56
& 15.43
& 1.87
& (1) / 0
& 10.51 \\

Qwen3-235BA22B~\citep{yang2025qwen3}
& 15.50
& 5.71
& 13.56
& 4.03
& (0) / 0
& 9.70 \\

\midrule
\multicolumn{7}{c}{\cellcolor{CornflowerBlue!15}\textit{\textbf{SearchSWE}}} \\ 
\midrule
GLM-4.7~\citep{zai2025glm47}
& \textbf{45.40}\DeltaNotePlus{+5.2}
& \underline{32.39}\DeltaNote{-3.7}
& \textbf{39.77}\DeltaNote{-0.1}
& 49.44\DeltaNotePlus{+1.0}
& \underline{(3) / 1}
& \textbf{41.75}\DeltaNotePlus{+0.6} \\

DeepSeek-V3.2~\citep{liu2025deepseek}
& 39.49\DeltaNotePlus{+1.5}
& 31.88\DeltaNotePlus{+1.3}
& 34.09\DeltaNote{-2.4}
& \underline{53.64}\DeltaNote{-1.4}
& (4) / 0
& \underline{39.78}\DeltaNote{-0.2} \\

Kimi-K2~\citep{team2025kimi}
& \underline{39.90}\DeltaNotePlus{+2.9}
& \textbf{33.33}\DeltaNotePlus{+5.6}
& \underline{34.83}\DeltaNote{-4.7}
& 49.31\DeltaNote{-5.6}
& (2) / 1
& 39.34\DeltaNote{-0.5} \\

GPT-5.2~\citep{openai2025gpt52}
& 36.22\DeltaNotePlus{+3.2}
& 22.22\DeltaNote{-1.4}
& 33.90\DeltaNote{-0.4}
& \textbf{55.85}\DeltaNotePlus{+2.0}
& \textbf{(7) / 2}
& 37.05\DeltaNotePlus{+0.9} \\

Seed-Coder-1.8~\citep{seed2025seed}
& 36.92\DeltaNote{-5.0}
& 22.06\DeltaNotePlus{+3.5}
& 32.57\DeltaNotePlus{+1.1}
& 41.40\DeltaNote{-1.3}
& (1) / 0
& 33.24\DeltaNote{-0.4} \\

Qwen3-Coder-Plus~\citep{qwen3-coder-2025}
& 17.50\DeltaNote{-1.7}
& 17.14\DeltaNotePlus{+11.6}
& 16.28\DeltaNotePlus{+0.9}
& 1.38\DeltaNote{-0.5}
& (0) / 0
& 13.08\DeltaNotePlus{+2.6} \\

Qwen3-235BA22B~\citep{yang2025qwen3}
& 16.58\DeltaNotePlus{+1.1}
& 9.72\DeltaNotePlus{+4.0}
& 14.12\DeltaNotePlus{+0.6}
& 7.00\DeltaNotePlus{+3.0}
& (1) / 0
& 11.86\DeltaNotePlus{+2.2} \\

\bottomrule
\end{tabular}
}
\vspace{-0.5em}
\caption{
Evaluation results of other models on \benchmark{}.
This table includes \benchmark{} entries that are not reported in Table~\ref{tab:main_result}.
\textbf{Bold} and \underline{underlined} values indicate the best and second-best results within each OpenHands/SearchSWE block, respectively.
For SearchSWE rows, \textcolor{darkred}{red}/\textcolor{darkgreen}{green} deltas compare against the matched OpenHands result for the same model.
}
\label{tab:additional_results}
\end{table*}

\subsection{\benchmark{} Construction Details}
\label{app:beyondswe_construction}
This section expands the construction procedure described in Section~\ref{sec:bench}.
Table~\ref{tab:app_candidate_pipeline} summarizes the candidate sources, filtering stages, human verification, and final retained instances for the four tasks.
\begin{table*}[t]
\centering
\small
\begin{tabular}{@{}p{2.2cm}p{3.1cm}p{3.8cm}p{3.9cm}p{2.0cm}@{}}
\toprule
\textbf{Task} & \textbf{Candidate Source} & \textbf{Automatic Filtering} & \textbf{Human Verification} & \textbf{Final Instances} \\
\midrule
CrossRepo &
Merged pull requests in Python-dominant repositories containing external links &
Environment construction and stability inspection reduce about 3,000 candidates to about 800 executable candidates &
Review external-link relevance and check that rewritten issue statements preserve task context without solution-specific details &
200 issues / 67 repos \\
\midrule
DomainFix &
Pull requests from expert-selected repositories across 11 scientific and engineering fields &
Environment construction and stability inspection reduce about 800 candidates to about 200 executable candidates &
Three domain experts independently verify environmental correctness, domain complexity, and solution non-triviality &
72 issues / 12 repos \\
\midrule
DepMigrate &
Pull requests mentioning 23 widely used packages and relevant version upgrades &
LLM-based migration filtering plus environment construction reduce about 7,000 candidates to about 1,000 executable candidates &
Four software engineering experts verify that each instance is a genuine dependency migration &
178 issues / 120 repos \\
\midrule
Doc2Repo &
Recently created Python repositories with continued activity, at least three contributors, and more than 20 stars &
Repository masking, specification generation, test adaptation, and executable environment construction produce 60 candidates &
Manual review selects high-quality specifications and checks alignment between adapted tests and documented behavior &
50 repos \\
\bottomrule
\end{tabular}
\caption{Candidate collection, filtering, and verification pipeline for \benchmark{}.}
\label{tab:app_candidate_pipeline}
\end{table*}

\textbf{Candidate collection.}
For CrossRepo, DomainFix, and DepMigrate, candidates are collected from merged pull requests so that each instance is grounded in a real developer change and has an executable reference patch.
CrossRepo starts from Python-dominant repositories whose pull requests contain external links; the links serve as evidence that developers used or referenced information outside the target repository.
DomainFix starts from repositories selected with domain experts across scientific and engineering fields, then collects pull requests from those repositories.
DepMigrate starts from pull requests mentioning major version changes of widely used dependencies.
Doc2Repo differs from the repair tasks: it starts from recently created Python repositories and converts each repository into a clean-room specification and test suite.

\textbf{Filtering and problem statements.}
The filtering process removes candidates that are not executable, not stable, or not aligned with the intended task category.
For issue-resolution tasks, raw pull requests often contain implementation details, commit references, or code-level hints.
We therefore rewrite them into issue-style problem statements that preserve the observable task requirements while removing solution-specific details.
For CrossRepo, reviewers additionally check that the external links are relevant to the issue rather than incidental references.
For Doc2Repo, repository names are masked as \texttt{target\_repo}; the specification preserves public API and behavioral requirements but removes implementation details and explicit directory structure.

\textbf{Retention criteria.}
An instance is retained only if its environment is reproducible and its tests distinguish the buggy state from the fixed state.
For CrossRepo, DomainFix, and DepMigrate, this requires P2P tests to pass and F2P tests to fail before applying the reference patch, and all selected tests to pass after applying the reference patch.
For Doc2Repo, the generated specification and adapted tests are checked against the reference implementation and manually audited for consistency.

\subsection{Human Verification and Quality Control}
\label{app:human_verification}

Table~\ref{tab:app_human_verification} summarizes the human verification protocol.
\begin{table*}[t]
\centering
\small
\begin{tabular}{@{}p{2.2cm}p{3.2cm}p{5.2cm}p{4.0cm}@{}}
\toprule
\textbf{Task} & \textbf{Reviewers} & \textbf{Verification Criteria} & \textbf{Agreement / Retention Rule} \\
\midrule
CrossRepo &
Senior software engineers and senior PhD auditors &
External links are relevant to the issue; rewritten statements preserve task requirements; no solution-specific details are exposed; tests exhibit valid fail-to-pass behavior &
Retain only instances passing manual relevance, leakage, and stability checks \\
\midrule
DomainFix &
Three domain experts, plus final senior PhD audit &
Environment executes as expected; the issue requires domain-specific knowledge beyond general programming; the solution is non-trivial and cannot be inferred from error messages alone &
Retain only instances accepted by all three domain experts \\
\midrule
DepMigrate &
Four software engineering experts, plus final senior PhD audit &
The pull request corresponds to a real dependency migration; the upgraded dependency is installed in the evaluation environment; the required fix is not an incidental local patch &
Retain only instances verified as genuine migration tasks \\
\midrule
Doc2Repo &
Senior software engineers and senior PhD auditors &
Specification preserves public API and behavior while removing implementation details; adapted tests match the specification; stricter-than-source requirements are documented &
Retain only specifications and tests judged aligned and executable \\
\midrule
All tasks &
Five senior software engineers and five senior PhD researchers &
Environment construction, data cleaning, problem-statement quality, test validity, and final task-category fit &
Reject instances with instability, solution leakage, underspecified behavior, or task-category mismatch \\
\bottomrule
\end{tabular}
\caption{Human verification protocol for \benchmark{}.}
\label{tab:app_human_verification}
\end{table*}

Human verification serves two purposes.
First, it checks whether a retained instance reflects the intended capability rather than an artifact of data collection, environment setup, or test construction.
Second, it checks whether the problem statement and test suite are fair: the task should provide enough information to identify the required behavior, but should not reveal the gold solution.

\textbf{Review protocol.}
DomainFix uses domain experts because the core question is whether the issue requires specialized scientific or engineering knowledge.
DepMigrate uses software engineering experts because the central question is whether the pull request represents a genuine dependency migration rather than an incidental version update.
Across all tasks, senior software engineers inspect environment construction and data cleaning, while senior PhD researchers in software engineering and LLMs audit final task validity.
Instances are rejected when reviewers identify unstable tests, insufficient task information, solution leakage, mismatch between the problem statement and reference patch, or a task-category mismatch.

\textbf{Compensation policy.}
Some reviewers and experts were members of the author team and participated as part of the research process.
External reviewers and experts were compensated according to the applicable project and institutional policies for their review roles.
We report this at the policy level rather than listing payment amounts, since compensation depends on the review arrangement and task batch rather than on a single uniform rate.

\section{Representative Task Examples}
\label{app:task_examples}

This section provides compact examples of the four task formats in \benchmark{}.
Each example highlights what information is given to the agent, what capability the instance stresses, and how the output is evaluated.

\begin{figure*}[t]
\begin{examplebox}{CrossRepo Example: \texttt{kitware\_trame-server\_pr8}}
\textbf{Task goal.} Fix an issue where the server ignores the explicit \texttt{host} argument and the \texttt{TRAME\_DEFAULT\_HOST} environment variable, causing it to bind to \texttt{localhost} even when downstream integrations require another interface.

\textbf{Capability stressed.} The issue is linked to downstream PyVista usage, so the agent must reason beyond the target repository and understand how server binding behavior affects related projects.

\textbf{Input signal.} The problem statement provides a reproduction script, expected binding behavior, and an external integration reference, but not the implementation location of the fix.

\textbf{Evaluation signal.} A valid patch must preserve existing behavior while making the host argument and environment-variable override pass the target F2P tests.
\end{examplebox}
\caption{Representative CrossRepo instance.}
\end{figure*}

\begin{figure*}[t]
\begin{examplebox}{DomainFix Example: \texttt{cvxpy\_cvxpy\_pr2125}}
\textbf{Task goal.} Add sparse Cholesky decomposition support for positive definite sparse matrices in \texttt{cvxpy.utilities.linalg}.

\textbf{Capability stressed.} The task is not only an API addition: the agent must understand the mathematical contract of sparse Cholesky factorization, including the permutation relation \(PAP^T = LL^T\), and implement it in a way compatible with sparse matrix representations.

\textbf{Input signal.} The problem statement gives the desired function behavior and example usage, but the implementation requires domain knowledge about numerical linear algebra and sparse factorization.

\textbf{Evaluation signal.} A valid patch must return the expected sparse triangular factor and permutation behavior while preserving existing CVXPY linear-algebra functionality.
\end{examplebox}
\caption{Representative DomainFix instance.}
\label{fig:example_domainfix}
\end{figure*}

\begin{figure*}[t]
\begin{examplebox}{DepMigrate Example: \texttt{stanfordnlp\_dsp\_pr403}}
\textbf{Task goal.} Update an LM client to support \texttt{openai>=1.0}, including Azure OpenAI configurations where \texttt{model}, \texttt{engine}, and \texttt{deployment\_id} handling changed.

\textbf{Capability stressed.} The agent must perform migration reasoning rather than a localized bug fix: it needs to map old SDK assumptions to the new client interface, handle Azure-specific behavior, and avoid breaking existing cache compatibility.

\textbf{Input signal.} The issue states current failure modes under the upgraded dependency and lists compatibility constraints, but the agent must inspect the repository to find all affected client initialization and inference paths.

\textbf{Evaluation signal.} A valid patch must pass tests under the upgraded dependency while preserving behavior for standard OpenAI and Azure-backed configurations.
\end{examplebox}
\caption{Representative DepMigrate instance.}
\label{fig:example_depmigrate}
\end{figure*}

\begin{figure*}[t]
\begin{examplebox}{Doc2Repo Example: \texttt{doc2mark}}
\textbf{Task goal.} Construct a repository from a specification for a unified document-processing library that converts PDFs, Office files, images, HTML, and related formats into Markdown.

\textbf{Capability stressed.} The agent starts from an empty workspace and must infer the repository structure, public API, dependencies, and implementation behavior from the specification alone.

\textbf{Input signal.} The specification describes classes such as \texttt{UnifiedDocumentLoader}, method signatures, supported formats, OCR configuration, caching behavior, and expected exceptions, while masking the original repository identity as \texttt{target\_repo}.

\textbf{Evaluation signal.} A valid submission must implement a coherent package whose public API and behavior pass the adapted repository test suite.
\end{examplebox}
\caption{Representative Doc2Repo instance.}
\label{fig:example_doc2repo}
\end{figure*}

\section{Prompt Design and Excerpts}
\label{app:prompt_design}

This section summarizes the prompts used for benchmark construction and search-augmented coding.
We avoid printing the full prompts in the paper because they are long and mostly operational.
Instead, Table~\ref{tab:prompt_design_summary} summarizes their roles, and the boxes below show the key constraints needed to understand the evaluation design.
\begin{table*}[t]
\centering
\small
\resizebox{\textwidth}{!}{
\begin{tabular}{@{}p{3.0cm}p{4.2cm}p{5.2cm}p{2.4cm}@{}}
\toprule
\textbf{Prompt Component} & \textbf{Purpose} & \textbf{Key Constraints} & \textbf{Used For} \\
\midrule
Problem-statement generation &
Convert raw pull requests into issue-style task descriptions &
Remove solution-specific details while preserving reproduction steps, expected behavior, logs, and necessary context &
CrossRepo, DomainFix, DepMigrate \\
\midrule
\agent{} system prompt &
Define the shared code-agent workflow with search and local verification &
Search only when local context is insufficient; fetch full pages before using results; prioritize local environment and user instructions &
All \agent{} runs \\
\midrule
CrossRepo user prompt &
Encourage use of external software artifacts when target-repository context is incomplete &
Inspect related projects or documentation, but adapt evidence to the local repository and tests &
CrossRepo \\
\midrule
DomainFix user prompt &
Guide agents to combine code reasoning with domain-specific knowledge &
Verify scientific definitions, formulas, or domain conventions before applying a fix &
DomainFix \\
\midrule
DepMigrate user prompt &
Frame the task as compatibility with upgraded dependencies &
Do not downgrade dependencies; use migration guides or release notes only when compatible with installed versions &
DepMigrate \\
\midrule
Doc2Repo user prompt &
Constrain clean-room repository construction from a specification &
Treat the provided specification as authoritative; use search only for dependency syntax or missing imports, not to override the spec &
Doc2Repo \\
\bottomrule
\end{tabular}
}
\caption{Summary of prompt components used for benchmark construction and \agent{} evaluation.}
\label{tab:prompt_design_summary}
\end{table*}

\subsection{Problem Statement Generation}

For CrossRepo, DomainFix, and DepMigrate, raw pull-request descriptions may contain implementation details, commit references, or direct descriptions of the fix.
We therefore use a problem-statement generation prompt to rewrite pull requests into issue-style task descriptions.
The prompt is designed to preserve observable task context while removing solution-specific information.

\begin{figure*}[t]
\begin{promptbox}{Problem Statement Generation Prompt: Key Constraints}
You are an expert-level autonomous software engineer and open-source maintainer.
Your sole task is to draft a concise, human-like GitHub issue based on a provided pull request.

Critical rules:
- Do not reveal the solution, diff, internal function names, line numbers, PR author, commit hash, or fix logic.
- Write from the perspective of a user or developer reporting the problem before the fix exists.
- Preserve reproduction steps, observed behavior, expected behavior, error logs, environment details, and external context needed to make the task solvable.
- The reproduction should be a natural script or command sequence, not a unit test that asserts the answer.
\end{promptbox}
\caption{Excerpt of the prompt used to convert pull requests into issue-style problem statements.}
\label{fig:prompt_problem_statement_excerpt}
\end{figure*}

\subsection{\agent{} Prompt Design}

\agent{} uses a shared system prompt plus task-specific user prompts.
The shared system prompt defines the local coding workflow, search-use policy, and verification discipline.
Task-specific prompts adjust the mission context: CrossRepo encourages external software investigation; DomainFix highlights domain knowledge; DepMigrate emphasizes compatibility with upgraded dependencies; Doc2Repo treats the provided specification as the authoritative source.

\begin{figure*}[t]
\begin{promptbox}{\agent{} System Prompt: Search-Use Policy Excerpt}
Use search strategically, not exclusively.
Search only when the task requires information that is missing from the local repository.

When to search:
- Unknown syntax or API details of a specific library version.
- Obscure error messages that cannot be diagnosed from local code.
- Missing implementation details not provided in the local context.

Search methodology:
- Use focused queries to identify relevant sources.
- If a result looks promising, fetch and read the actual page content.
- Never write code based only on search snippets.
- Before using an online library pattern, check whether the library is installed locally and verify the local version/API.
- User instructions, local files, and provided specifications are the highest-priority constraints.
\end{promptbox}
\caption{Excerpt of the shared \agent{} system prompt governing search use and local verification.}
\label{fig:prompt_search_policy_excerpt}
\end{figure*}

\subsection{Task-Specific Prompt Roles}

The task-specific user prompts differ mainly in the constraints they emphasize.
For CrossRepo, the prompt asks the agent to inspect external artifacts when local context is insufficient.
For DomainFix, it encourages the agent to verify scientific definitions, formulas, or domain conventions before modifying code.
For DepMigrate, it forbids dependency downgrades and requires compatibility with the upgraded package versions installed in the environment.
For Doc2Repo, it treats the provided specification as the ground truth and restricts search to missing dependencies or library-version syntax, rather than architectural invention.

These prompts are used to make search behavior explicit and auditable.
They are not intended as a new agent architecture; they define a controlled way to test whether agents can combine external evidence with repository-local coding and verification.

\end{document}